\documentclass[letterpaper]{article}
\usepackage{aaai2027}
\usepackage[hyphens]{url}  
\usepackage{graphicx} 
\usepackage{natbib}  
\usepackage{caption} 
\usepackage{booktabs}
\usepackage{amsmath}
\usepackage{amssymb}
\usepackage{multirow}
\usepackage{fvextra}
\usepackage{tabularx}
\usepackage{array}

\newcommand{\cmark}{\ensuremath{\checkmark}}
\newcommand{\xmark}{\ensuremath{\times}}
\DefineVerbatimEnvironment{myverb}{Verbatim}{
  breaklines=true,
  breakanywhere=true,
  breaksymbolleft={},
  breaksymbolright={},
  breaksymbolsepleft=0pt,
  breaksymbolsepright=0pt,
  breaksymbolindentleft=0pt,
  breaksymbolindentright=0pt,
  breakindent=0pt,
  breakautoindent=false,
  breakanywheresymbolpre={},
  breakanywheresymbolpost={},
  formatcom=\normalfont,
  fontsize=\normalsize
}

\title{MolGVR: A Chemistry-Grounded Framework for Text-to-Molecule Generation}
\author{Qian Tan\textsuperscript{1}, 
Xuanyu Zhu\textsuperscript{3}, 
Lei Jiang\textsuperscript{1}, 
Zhonghang Yuan\textsuperscript{1},\\
\textbf{Chen Zhang\textsuperscript{2},
Yuqiang Li\textsuperscript{2,\dag},}}
\affiliations{\textsuperscript{1}University of Science and Technology of China \\
\textsuperscript{2}Shanghai Artificial Intelligence Laboratory \\
\textsuperscript{3}Shanghai Jiao Tong University \\
\texttt{tanqian@mail.ustc.edu.cn,liyuqiang@pjlab.org.cn} }

\begin{document}

\nocopyright
\maketitle

\begin{abstract}
Text-to-molecule generation is typically formulated as a one-shot sequence generation problem, where a model directly maps target descriptions to molecular representations. However, molecular descriptions often contain informative structural constraints, and violating such constraints can change the molecular identity. This makes chemical verification and error correction important but underexplored. To fill this gap, we propose MolGVR, a chemistry-grounded Generator--Verifier--Refiner framework. The Generator infers structural evidence and generates candidate molecules. The Verifier addresses the lack of chemical validation by converting descriptions into chemical constraints and checking candidates against them. The Refiner addresses generation failures by revising candidates rejected by the Verifier. Experiments on ChEBI-20 and PCDes show that MolGVR improves exact-match performance. These results suggest that coupling generation with executable verification and feedback-guided refinement is an effective way to improve text-to-molecule generation.
\end{abstract}

\section{Introduction}
\label{sec:intro}

Translating natural language descriptions into molecular representations has emerged as a core problem in molecule-language modeling~\cite{edwards2021text2mol}. This task requires a model to understand the structural semantics in chemical descriptions while producing a valid molecular representation that is aligned with the description. MolT5~\cite{edwards2022translation} formulated the molecule-caption translation task, and a growing body of work has advanced text-to-molecule generation from multiple perspectives~\cite{liu2023molca,pei2025dmolt,li2024empowering,jang2025structural,wang2025chem}.
Sequence-to-sequence methods represented by MolT5~\cite{edwards2022translation} learn direct mappings between molecular strings and natural language; ICMA~\cite{li2025large} improve generation with retrieved contexts, example reranking, and in-context molecule learning; and MSR~\cite{jang2025structural} improves molecular generation performance by introducing intermediate reasoning steps.
Together, these methods have substantially improved text-to-molecule generation performance.

\begin{figure*}[t]
\centering
\includegraphics[width=14cm]{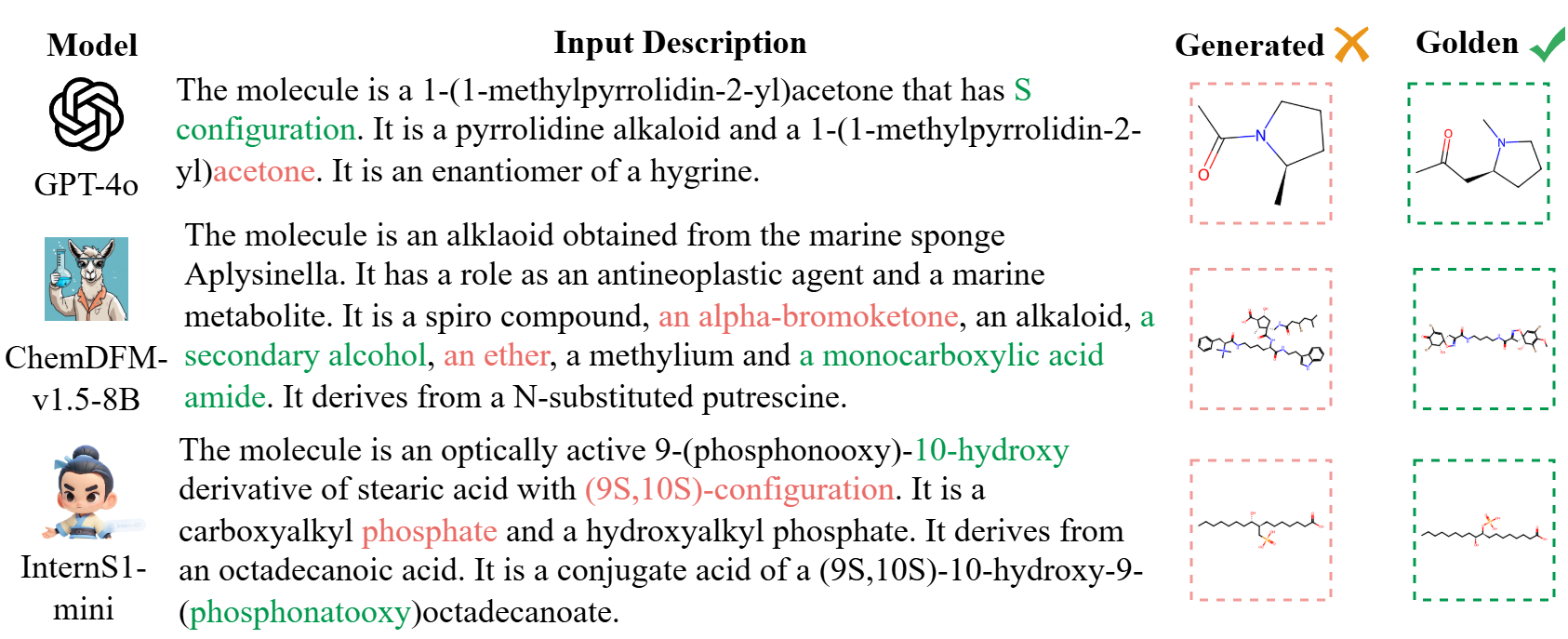}

\caption{Motivation of MolGVR. Existing general-purpose, chemistry, and scientific LLMs can generate molecules that appear plausible, yet still violate explicit chemical constraints in the text. Red highlights indicate violated descriptions, while green highlights indicate satisfied constraints.}
\label{fig:motivation}
\end{figure*}

Despite this progress, current text-to-molecule methods are still dominated by a one-shot prediction paradigm: they improve generation performance by learning stronger mappings between textual descriptions and molecular representations~\cite{edwards2022translation,li2024empowering,li2025large,jang2025structural}.
However, we observe a recurring failure mode of this paradigm: generated outputs may appear plausible while being either invalid SMILES or inconsistent with explicit constraints in the target description. We refer to this limitation as the verification-and-correction gap. As illustrated in Figure~\ref{fig:motivation}, this recurring noncompliance pattern appears across representative models: GPT-4o~\cite{hurst2024gpt4o} as a general-purpose LLM, ChemDFM~\cite{zhao2024chemdfm} as a chemistry foundation model, and Intern-S1-mini~\cite{bai2025intern} as a multidisciplinary scientific foundation model.
For instance, in the ChemDFM-v1.5-8B example, the description explicitly requires an ether group, but the generated molecule lacks the corresponding substructure. These errors indicate that one-shot generation does not fully exploit the chemistry-grounded information contained in the target description.
As a result, even clear and checkable constraint violations may remain uncorrected, limiting the structural fidelity of text-to-molecule generation. This motivates a more process-oriented view of text-to-molecule generation.
In real-world molecular design, candidate molecules are rarely accepted in a single step; instead, they are proposed, evaluated against target requirements, and revised according to feedback~\cite{plowright2012hypothesis}.
Recent systems such as ChatMol also indicate a broader trend toward interactive and agentic molecular design~\cite{zeng2024chatmol}.
These observations suggest that text-to-molecule generation can also benefit from moving beyond one-shot prediction.

To this end, we propose MolGVR, a chemistry-grounded Generator--Verifier--Refiner framework for text-to-molecule generation.
The Generator infers chemistry-grounded evidence and produces candidate molecules.
The Verifier addresses the lack of chemical validation by converting description constraints into executable chemical checks and validating candidate molecules against them.
The Refiner uses verifier-provided failure reasons to perform targeted correction on rejected candidates.
Through this workflow, MolGVR turns text-to-molecule generation from one-shot prediction into a chemistry-grounded process in which specialized components collaborate to detect and refine constraint-level errors.
This design is aligned with practical molecular design workflows and provides a step toward more application-oriented molecule generation systems.

Our contributions can be summarized as follows:
\begin{itemize}
    \item We propose MolGVR, a chemistry-grounded Generator--Verifier--Refiner framework that mitigates the limitation of one-shot text-to-molecule generation. By integrating executable verification and targeted refinement, MolGVR makes text-to-molecule generation more faithful.

    \item We demonstrate strong performance on two text-to-molecule benchmarks (0.582 match score on ChEBI-20 and 0.500 on PCDes), showing advantages in exact molecular agreement over strong one-shot baselines.

    \item We show through ablation analyses that MolGVR's gains arise from verification-guided refinement rather than repeated generation, and Verifier--Refiner can be combined with multi-sample to further improve performance.
\end{itemize}

\section{Related Work}
\begin{figure*}[!t]
\centering
\includegraphics[width=14cm]{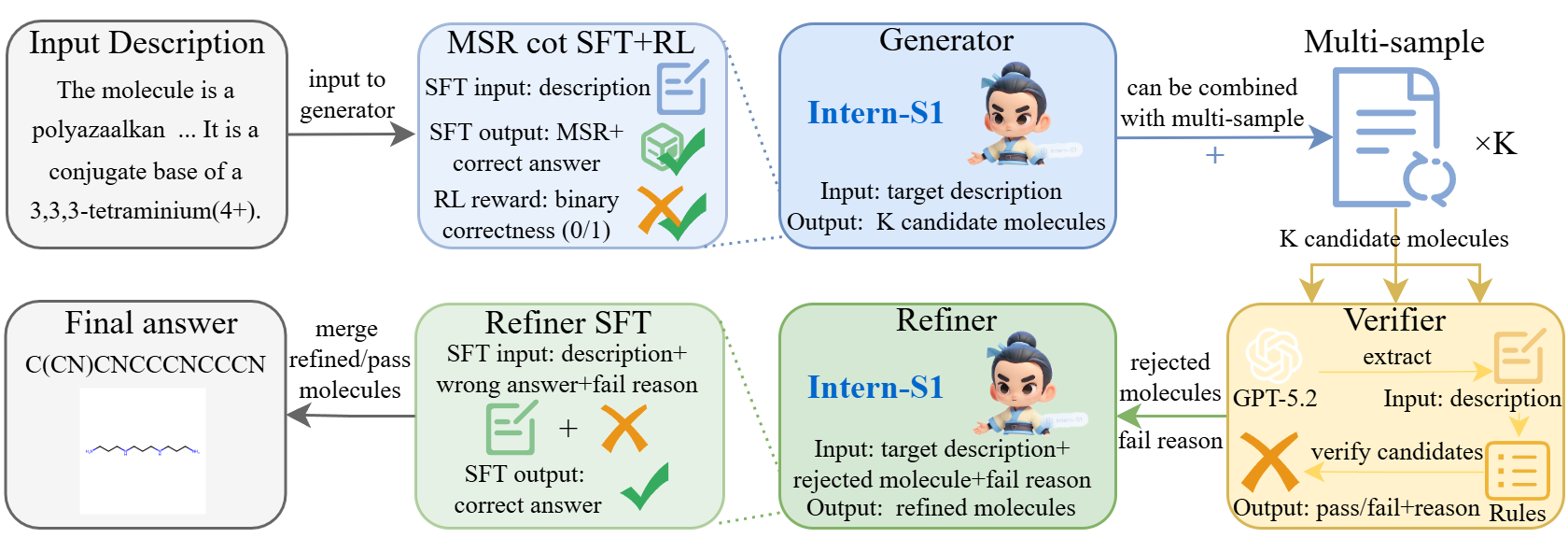}

\caption{
Overview of the MolGVR workflow. The MSR- and RL-trained Generator produces one or multiple candidate molecules from an input description. The Verifier extracts chemical rules, checks each candidate, and returns pass/fail decisions with failure reasons. Rejected candidates are refined using the description, molecule, and failure reason, then merged with accepted candidates to form the final output.
}
\label{fig:framework}
\end{figure*}

\paragraph{Text-to-molecule generation.}
Text2Mol~\cite{edwards2021text2mol}, MolT5~\cite{edwards2022translation}, and PCDes~\cite{zeng2022deep} established molecule--text alignment and benchmarked generation from natural-language descriptions. Subsequent work improved generators through structural modeling and chemistry-specific pretraining~\cite{liu2023molca,pei2025dmolt,yu2024llasmol,zhang2024chemllm,zhao2024chemdfm,deng2025chemical}, diffusion or latent-space generation~\cite{gong2024text,chang2025ldmol}, improved tokenization~\cite{kim-etal-2025-training}, and retrieval or fine-grained cross-modal alignment~\cite{li2025large,li2026molreflect}. These methods primarily focus on improving the generator itself.

\paragraph{Structural reasoning and training-time enhancement.}
MSR~\cite{jang2025structural} introduces intermediate structural reasoning, while Chem-R~\cite{wang2025chem} and MolReasoner~\cite{zhao2025molreasoner} combine chemistry-aware reasoning supervision with reinforcement learning. Self-augmentation~\cite{jiang2024enhancing} and large-scale instruction tuning~\cite{yu2024llasmol} further improve generation through stronger supervision. These approaches enhance model training but do not explicitly verify and correct generated molecules.

\paragraph{Multi-agent and agentic molecular design.}
Agentic methods apply iterative reasoning, tools, and role specialization to molecular tasks. MotifAgent~\cite{feng2026motifagent} learns motif-level assembly through multiple agents, while ChatMol~\cite{zeng2024chatmol}, DrugAssist~\cite{ye2025drugassist}, AgentDrug~\cite{le-etal-2025-agentdrug}, MT-MOL~\cite{kim-etal-2025-mt}, and MADD~\cite{solovev2025madd} target conversational design, molecular optimization, drug editing, or hit discovery. Their focus differs from verification-guided text-to-molecule generation.

\section{Method}
\noindent\textbf{Problem Formulation and Overview}.
Let $\mathcal{D}=\{(x_i,m_i^\star)\}_{i=1}^{N}$ denote a dataset, where $x_i$ is a target description and $m_i^\star$ is the corresponding molecule represented as a SMILES~\cite{weininger1988smiles}\footnote{A SMILES (Simplified Molecular Input Line Entry System) string is a compact, text-based representation of a molecule's structure that encodes its atomic composition and connectivity in a linear format. }.
The goal is to generate a molecule $\hat{m}_i$ that is structurally faithful to $x_i$, ideally matching $m_i^\star$ under molecule-level equivalence.

Conventional methods typically learn a direct generator
\begin{equation}
    \hat{m}_i \sim p_{\theta}(\cdot \mid x_i),
\end{equation}
which maps the description to a molecular string in a one-shot manner.
However, as discussed in the Introduction, such direct generation may produce molecules that appear plausible but invalid or violate explicit structural constraints in the description.
MolGVR mitigates this verification-and-correction gap by decomposing the prediction process into three coordinated stages: generation, verification, and refinement, as illustrated in Figure~\ref{fig:framework}.

First, the Generator provides a chemistry-grounded initial prediction.
Given a description $x_i$, it produces intermediate structural evidence $c_i$  and generates one or more candidate molecules:
\begin{equation}
    \mathcal{C}_i = \{(c_{i,j}, \hat{m}_{i,j})\}_{j=1}^{K},
    \quad (c_{i,j}, \hat{m}_{i,j}) \sim G_{\theta}(\cdot \mid x_i),
\end{equation}
where $K$ is the number of sampled candidates.
After that, the Verifier exploits description-derived constraints for explicit error detection.
It extracts a set of checkable rules $R_i$ from the description and executes them against candidate molecules:
\begin{equation}
    R_i = \mathcal{E}(x_i),
    \quad
    \mathcal{V}(R_i, \hat{m}_{i,j})
    \rightarrow
    \{\texttt{pass}, (\texttt{fail}, f_{i,j})\},
\end{equation}
where $\mathcal{E}(\cdot)$ denotes rule extraction, $\mathcal{V}(\cdot)$ denotes tool-based verification, and $f_{i,j}$ denotes the failure reason.
Candidates that pass are retained, while candidates that fail are passed to the Refiner together with their failure reasons.
Finally, the Refiner performs correction on verifier-rejected candidates:
\begin{equation}
\tilde{m}_{i,j}
\sim
F_\phi(\cdot\mid x_i,\hat{m}_{i,j},f_{i,j}),
    \quad \text{if } \texttt{fail},
\end{equation}
where $F_{\phi}$ denotes the Refiner.
The final candidate set is therefore
\begin{equation}
    \mathcal{M}_i =
    \{\hat{m}_{i,j}\mid \texttt{pass}\}
    \cup
    \{\tilde{m}_{i,j}\mid\texttt{fail}\}.
\end{equation}

By integrating the three components, MolGVR utilizes target information for post-generation error detection and correction, thereby improving molecular fidelity and mitigating the verification-and-correction gap.

\subsection{MolGVR Generator}
\label{sec:gen}

\begin{table*}[t]
\centering
\small
\setlength{\tabcolsep}{2pt}
\begin{tabular}{
@{}
>{\raggedright\arraybackslash}p{0.08\linewidth}
@{\hspace{2pt}}
>{\raggedright\arraybackslash}p{0.16\linewidth}
@{\hspace{2pt}}
>{\raggedright\arraybackslash}p{0.22\linewidth}
@{\hspace{8pt}}
>{\raggedright\arraybackslash}p{0.08\linewidth}
@{\hspace{2pt}}
>{\raggedright\arraybackslash}p{0.17\linewidth}
@{\hspace{2pt}}
>{\raggedright\arraybackslash}p{0.26\linewidth}
@{}
}
\toprule
Constraint
&
Example JSON Rule
&
Deterministic Verification
&
Constraint
&
Example JSON Rule
&
Deterministic Verification \\
\midrule

Invalid SMILES
&
--
&
Parse with Chem.MolFromSmiles; returning None indicates a violation.
&
Core scaffold / ring
&
\{"text\_span": "benzene",
"normalized": "benzene"\}
&
Compile the benzene SMARTS
c1ccccc1 and test its presence with
mol.HasSubstructMatch. \\

\addlinespace[2pt]

Functional group
&
\{"text\_span":
"haloacetic acid",
\par
"normalized":
"carboxylic\_acid"\}
&
Match the carboxylic-acid SMARTS
[CX3]\allowbreak(=O)[OX2H1] using
Chem.Mol\allowbreak
FromSmarts and
mol.HasSubstructMatch.
&
Salt / charge state
&
\{"text\_span":
"zwitterionic form",
"normalized":
"zwitterion"\}
&
Identify zwitterions by requiring both positive and negative atom charges via atom.GetFormalCharge. \\

\addlinespace[2pt]

Element count
&
\{"text\_span":
"an iodine atom",
"normalized": "iodine",
"count": 1\}
&
Count iodine atoms with atom.GetSymbol and match the extracted count.
&
Element / atom type
&
\{"text\_span":
"iodine atom",
"normalized": "iodine"\}
&
Map iodine to the atomic symbol I.
Count atom symbols using atom.GetSymbol
and require at least one iodine atom. \\

\addlinespace[2pt]

Group count
&
\{"text\_span": "two carboxy groups",
"normalized": "carboxylic\_acid",
"count": 2\}
&
Count unique matches of
[CX3](=O)\allowbreak[OX2H1] and
[CX3](=O)\allowbreak[O-] using
mol.\allowbreak GetSubstructMatches(...,\allowbreak
uniquify=True).
&
Stereo
\par
chemistry
&
\{"text\_span": "S configuration",
"normalized": "s\_center",
"count": 1\}
&
Assign stereochemistry and count S centers with Chem.AssignStereochemistry and Chem.FindMolChiralCenters. \\

\bottomrule
\end{tabular}

\caption{Representative examples of the eight constraint types and their
deterministic implementations. The invalid-SMILES rule is built into the
Verifier, while the remaining rules are extracted from molecular descriptions.}
\label{tab:rule_checker_examples}
\end{table*}

The MolGVR Generator is built on Intern-S1-mini~\cite{bai2025intern} and post-trained on the training splits of ChEBI-20~\cite{edwards2021text2mol,edwards2022translation} and PCDes~\cite{zeng2022deep}. Its training consists of two stages: supervised fine-tuning (SFT) and reinforcement learning (RL).

\subsubsection{Generator SFT}

Following MSR~\cite{jang2025structural}, we use RDKit to extract six types of structural evidence from each gold SMILES: (1) molecular formula, (2) longest carbon chain length, (3) aromatic ring information, (4) ring information, (5) functional groups, and (6) chirality. As shown in Figure~\ref{fig:data}(a) in Appendix~\ref{sec:app_data_synthesis}, these attributes are organized into a tagged evidence sequence:
\begin{equation}
\begin{aligned}
c^\star = [&\texttt{<mol\_formula>} \cdots \texttt{</mol\_formula>}, \\
          &\ldots, \\
          &\texttt{<chiral>} \cdots \texttt{</chiral>}].
\end{aligned}
\end{equation}
The supervised target is then formed by concatenating the evidence sequence with the gold molecule:
\begin{equation}
y^\star = [c^\star; m^\star].
\end{equation}

We fine-tune the generator with the standard SFT objective. The loss is
\begin{equation}
\mathcal{L}_{\mathrm{SFT}}^{\mathrm{gen}}(\theta)
=
-\sum_{i=1}^{N}\sum_{t=1}^{|y_i^\star|}
\log G_{\theta}\!\left(y_{i,t}^\star \mid x_i, y_{i,<t}^\star\right),
\label{eq:gen_sft}
\end{equation}
where $y_{i,t}^\star$ is the $t$-th target token.

\subsubsection{Generator RL}

Starting from the SFT checkpoint, we apply GRPO~\cite{shao2024deepseekmath} with a format reward $r_{\mathrm{fmt}}$ that encourages the six reasoning tags and boxed output, and a binary exact-match accuracy reward $r_{\mathrm{acc}}$ computed from the boxed answer. To avoid repeated RDKit canonicalization during rollouts, accuracy is evaluated by raw-string matching against the gold answer. The final reward is
$(1-\lambda)r_{\mathrm{acc}}+\lambda r_{\mathrm{fmt}}$ with $\lambda=0.1$; detailed definitions are provided in Appendix~\ref{sec:app_reward_design}.

\subsection{MolGVR Verifier}
\label{sec:verify}

The Verifier consists of two stages: structured rule extraction where an LLM extracts checkable chemical constraints and deterministic rule execution against each generated candidate using RDKit-based checkers. 

\subsubsection{Rule Extraction}

Given a target description, we prompt GPT-5.2 to produce a JSON object following a predefined schema. Across the two datasets, the description-derived rule categories include functional groups, core scaffolds or ring systems, element or atom types, element counts, group counts, stereochemistry, and salt or charge states. Each dataset uses six description-derived categories together with invalid SMILES as a built-in seventh category that does not require LLM extraction. Representative JSON rules for these constraint types are shown in Table~\ref{tab:rule_checker_examples}. The detail schema and extraction prompt are provided in Appendix~\ref{sec:app_verify_rule}.

\subsubsection{Deterministic Rule Execution}

The extracted rules are translated by GPT-5.2 into RDKit-based checker code covering molecular parsing, SMARTS substructure matching, molecular graph analysis, atom and group counting, stereochemistry assignment, and formal-charge inspection. The resulting checkers are then executed deterministically on each candidate molecule. Table~\ref{tab:rule_checker_examples} provides representative mappings from the extracted JSON rules to their deterministic checker implementations. Each violation is recorded with its category and source text span. Additional implementation details are provided in Appendix~\ref{sec:app_verify_tool}.

To reduce false rejections caused by ambiguous or underspecified descriptions, the rule definitions and their corresponding checker logic are reviewed by chemistry experts for chemical validity and reliability. Only rules that pass this expert validation are allowed to affect the final pass--fail decision.

\subsection{MolGVR Refiner}
\label{sec:refine}

\begin{table*}[t]
\centering
\small
\setlength{\tabcolsep}{1.2pt}
\begin{tabular}{llcccccccc}
\toprule
Dataset & Model & Validity$\uparrow$ & BLEU$\uparrow$ & Levenshtein$\downarrow$ & MACCS FTS$\uparrow$ & RDK FTS$\uparrow$ & Morgan FTS$\uparrow$ & Match$\uparrow$ & FCD$\downarrow$ \\
\midrule
\multirow{19}{*}{ChEBI-20}
& Transformer~\cite{vaswani2017attention}                              & 0.906 & 0.499 & 57.660 & 0.480 & 0.320 & 0.217 & 0.000 & 11.32 \\
& GIT-Mol~\cite{liu2024git}              & 0.928 & 0.756 & 26.315 & 0.738 & 0.582 & 0.519 & 0.051 & - \\
& T5$_{\text{base}}$~\cite{raffel2020exploring} & 0.660 & 0.765 & 24.950 & 0.731 & 0.605 & 0.545 & 0.069 & 2.48 \\
& MolT5$_{\text{base}}$~\cite{edwards2022translation} & 0.772 & 0.769 & 24.458 & 0.721 & 0.588 & 0.529 & 0.081 & 2.18 \\
& T5$_{\text{large}}$                      & 0.902 & 0.854 & 16.721 & 0.823 & 0.731 & 0.670 & 0.279 & 1.22 \\
& MolT5$_{\text{large}}$                   & 0.905 & 0.854 & 16.071 & 0.834 & 0.746 & 0.684 & 0.311 & 1.20 \\
& MolXPT~\cite{liu2023molxpt}               & 0.983 & -     & -      & 0.859 & 0.757 & 0.667 & 0.215 & 0.45 \\
& bioT5~\cite{pei2023biot5}                 & \textbf{1.000} & 0.867 & 15.097 & 0.886 & 0.801 & 0.734 & 0.413 & \underline{0.43} \\
& bioT5+~\cite{pei2024biot5+}                & \textbf{1.000} & 0.872 & 12.776 & 0.907 & 0.835 & 0.779 & 0.522 & \textbf{0.35} \\
& MolReGPT~\cite{li2024empowering} & 0.900 & 0.860 & 17.140 & 0.900 & 0.800 & 0.740 & 0.280 & - \\
& Mol-Instructions~\cite{fang2024mol} & \textbf{1.000} & 0.300 & 39.420 & 0.440 & 0.290 & 0.250 & 0.040 & - \\
& MolReasoner~\cite{zhao2025molreasoner} & 0.970 & 0.780 & 26.930 & 0.680 & 0.440 & 0.360 & 0.080 & - \\
& Mol-R1~\cite{li2025mol} & 0.850 & 0.640 & 32.940 & 0.820 & 0.680 & 0.610 & 0.230 & - \\
& GPT-4o~\cite{hurst2024gpt4o} & 0.770 & 0.450 & 48.380 & 0.790 & 0.580 & 0.500 & 0.070 & - \\

& DeepSeek-R1~\cite{guo2025deepseek} & 0.780 & 0.510 & 169.360 & 0.920 & 0.820 & 0.750 & 0.220 & - \\
& ChemDFM-v1.5-8B~\cite{zhao2024chemdfm} & 0.977 & \textbf{0.903} & \textbf{11.782} & \underline{0.939} & \underline{0.880} & \underline{0.830} & \underline{0.530} & 0.84 \\
& ether0-24B~\cite{narayanan2026training} & 0.730 & 0.390 & 860.990 & 0.820 & 0.700 & 0.640 & 0.270 & - \\
& Chem-R-8B~\cite{wang2025chem} & 0.935 & 0.824 & 19.830 & 0.917 & 0.830 & 0.777 & 0.414 & 8.41 \\

& \textbf{MolGVR(ours)} & \underline{0.992} & \underline{0.886} & \underline{11.894} & \textbf{0.946} & \textbf{0.887} & \textbf{0.844} & \textbf{0.582} & 4.00\\

\midrule
\multirow{5}{*}{PCDes}
& MolT5$_{\text{large}}$ & 0.944 & 0.692 & 18.481 & \underline{0.810} & \underline{0.741} & \underline{0.699} & 0.440 & \underline{0.70} \\
& bioT5                  & \textbf{1.000} & \textbf{0.754} & \textbf{15.658} & 0.797 & 0.726 & 0.677 & \underline{0.455} & \textbf{0.69} \\
& bioT5+                 & \underline{0.999} & 0.677 & 20.464 & 0.743 & 0.615 & 0.541 & 0.266 & 1.09 \\
& Chem-R-8B & 0.941 & 0.649 & 24.721 & 0.808 & 0.716 & 0.654 & 0.411 & 0.74 \\

& \textbf{MolGVR (ours)} & 0.996 & \underline{0.743} & \underline{15.873} & \textbf{0.845} & \textbf{0.769} & \textbf{0.710} & \textbf{0.500} & 0.97\\

\bottomrule
\end{tabular}
\caption{Performance comparison on the ChEBI-20 and PCDes datasets.
MolGVR is evaluated under the pass@1 setting: the generator produces one candidate per input, verifier-rejected candidates are refined once, and the resulting candidates are merged for evaluation.
MolGVR achieves the best Match scores on both datasets, improving over the strongest listed baselines by \textbf{9.8\%} on ChEBI-20 (0.530$\rightarrow$0.582) and \textbf{9.9\%} on PCDes (0.455$\rightarrow$0.500), while also improving structural similarity metrics.
Best results are highlighted in bold and second-best results are underlined.}
\label{tab:main_results}
\end{table*}

Given a molecular description, a verifier-rejected candidate, and its corresponding failure reason, the Refiner is designed to revise the candidate into the correct structure through two SFT stages. In the first stage, the input consists of the description and incorrect candidate, while the target is the gold molecule. This stage develops a general refinement capability. In the second stage, the Refiner is further trained with failure-aware inputs consisting of the description, rejected molecule, and corresponding failure reason.

\subsubsection{Refiner SFT}

As illustrated in Figure~\ref{fig:data}(b) in Appendix~\ref{sec:app_data_synthesis}, we sample $K=20$ candidates for each training input from the trained Generator, canonicalize each candidate and the gold molecule, and retain non-matching candidates. For the first stage, each retained candidate is paired with the original description and gold molecule to form an input--target instance. For the second stage, each retained candidate is further examined by the Verifier, checks whether the candidate satisfies structural constraints from the description. Candidates that fail the verification are paired with their corresponding failure reasons, forming refinement instances whose input consists of the description, rejected molecule, and failure reason, while the target remains the gold molecule. We optimize the autoregressive objective in Eq.~(\ref{eq:gen_sft}), replacing $G_\theta$ and $y_i^\star$ with the Refiner $F_\phi$ and gold molecule $m_i^\star$. The input is
$z_i=[x_i;m_i^{\mathrm{fail}}]$ in stage one and
$z_i=[x_i;m_i^{\mathrm{fail}};f_i]$ in stage two, where $m_i^{\mathrm{fail}}$ is the rejected molecule.

\section{Experiment}
\subsection{Experimental Setup}

We evaluate MolGVR against a broad range of autoregressive baselines, including sequence-to-sequence models such as T5~\cite{raffel2020exploring}, molecular language models such as MolT5~\cite{edwards2022translation}, and general-purpose or chemistry LLMs such as GPT-4o~\cite{hurst2024gpt4o} and Chem-R-8B~\cite{wang2025chem}. These models generate molecular strings in a one-shot manner and  therefore serve as baselines for evaluating whether the proposed MolGVR framework improves molecule generation beyond standard one-pass decoding. Molecular strings are canonicalized with RDKit before metric computation. Outputs longer than 4,096 characters are treated as invalid molecules and excluded from all metrics except Validity and Match, where they remain in the denominator and are counted as failures. Further details on the datasets, implementation, inference procedure, and evaluation metrics are provided in Appendices~\ref{sec:app_data}, \ref{sec:app_implementation_details}, and \ref{sec:metrics}, respectively.

\subsection{Result on ChEBI-20}

Table~\ref{tab:main_results} shows that MolGVR achieves strong performance on ChEBI-20, particularly on exact matching and structure-level similarity metrics.
MolGVR obtains the highest Match score of 0.582, outperforming the strongest autoregressive baselines in the table, including ChemDFM-v1.5-8B (0.530) and bioT5+ (0.522).
It also achieves the best MACCS, RDK, and Morgan fingerprint similarities while maintaining a high validity of 0.992.
These results suggest that, compared with one-shot autoregressive methods, MolGVR improves target-level molecular correctness and structural consistency with the input descriptions.

Although MolGVR achieves the strongest performance on exact matching and fingerprint-based structural metrics, it does not dominate all evaluation dimensions.
For validity, MolGVR is slightly below several baselines that achieve a 1.0 score.
For BLEU and Levenshtein distance, MolGVR ranks second, only marginally behind ChemDFM-v1.5-8B. This discrepancy can be attributed to the fact that these metrics operate on the surface form of SMILES strings and are sensitive to token-level variations, even when the molecular structures are highly similar.
In addition, MolGVR underperforms bioT5+ in terms of FCD, which suggests that the current exact-match-oriented optimization improves more on instance-level molecular recovery.

Figure~\ref{fig:demo}(a) provides a qualitative example. The verifier extracts two constraints: a ketone-related rule from the phrase acetone, and a stereochemical rule requiring one s\_center from the phrase S configuration. The initial candidate preserves the ketone substructure but assigns the wrong chirality. The Verifier therefore returns a stereochemistry-requirement failure reason, which guides the Refiner to correct the chiral center and recover the target molecule. More examples can be found in Appendix~\ref{sec:more_exp_chebi20}.

\subsection{Result on PCDes}
\begin{figure}[!t]
\centering
\includegraphics[width=8cm]{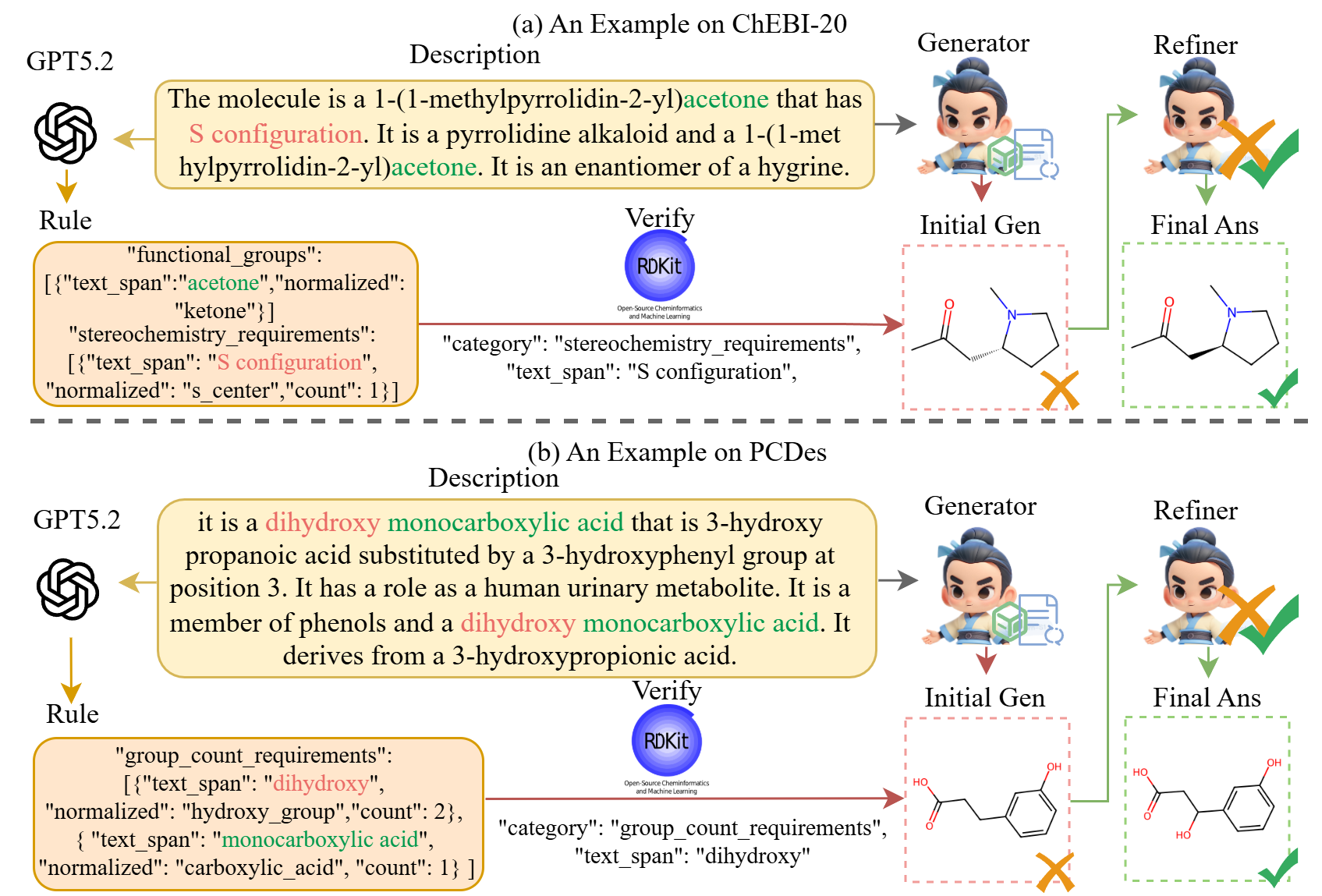}

\caption{Qualitative examples on the ChEBI-20 and PCDes datasets.
(a) The ChEBI-20 example shows a stereochemical violation detected by the Verifier and corrected using the corresponding failure feedback.
(b) The PCDes example shows a violated dihydroxy requirement caused by a missing side-chain hydroxyl group, which is identified by the Verifier and restored through feedback-guided refinement.}
\label{fig:demo}
\end{figure}

\begin{table*}[t]
\centering
\small
\setlength{\tabcolsep}{1mm}
\begin{tabular}{llcccccccc}
\toprule
Dataset & Method & Validity$\uparrow$ & BLEU$\uparrow$ & Levenshtein$\downarrow$ & MACCS FTS$\uparrow$ & RDK FTS$\uparrow$ & Morgan FTS$\uparrow$ & Match$\uparrow$ & FCD$\downarrow$ \\
\midrule
\multirow{6}{*}{ChEBI-20}
& Generator (pass@1) & 0.977 & 0.877 & 13.010 & 0.947 & 0.887 & 0.844 & 0.571 & 3.94 \\
& + Refiner (empty feedback) & 0.953 & 0.822 & 20.271 & 0.899 & 0.791 & 0.728 & 0.218 & \textbf{0.46} \\
& + Full pipeline & \underline{0.992} & 0.886 & 11.894 & 0.946 & 0.887 & 0.844 & 0.582 & 4.00 \\
& Generator (pass@5) & 0.991 & \underline{0.893} & \underline{11.563} & \underline{0.952} & \underline{0.898} & \underline{0.856} & \underline{0.593} & 2.44 \\
& + Refiner (empty feedback)& 0.967 & 0.846 & 18.159 & 0.906 & 0.801 & 0.741 & 0.228 & \underline{0.69} \\
& + Full pipeline & \textbf{0.998} & \textbf{0.904} & \textbf{10.339} & \textbf{0.954} & \textbf{0.901} & \textbf{0.861} & \textbf{0.605} & 2.48 \\
\midrule
\multirow{6}{*}{PCDes}
& Generator (pass@1) & 0.985 & 0.733 & 16.538 & 0.845 & 0.767 & 0.708 & 0.490 & \underline{0.97} \\
& + Refiner (empty feedback)& 0.958 & 0.566 & 33.763 & 0.788 & 0.665 & 0.591 & 0.213 & 53.83 \\
& + Full pipeline & \underline{0.996} & 0.743 & 15.873 & 0.845 & 0.769 & 0.710 & 0.500 & \underline{0.97} \\
& Generator (pass@5) & \underline{0.996} & \underline{0.756} & \underline{15.104} & \underline{0.858} & \underline{0.781} & \underline{0.725} & \underline{0.503} & \underline{0.97} \\
& + Refiner (empty feedback)& 0.972 & 0.722 & 20.482 & 0.799 & 0.677 & 0.607 & 0.221 & 53.02 \\
& + Full pipeline & \textbf{0.999} & \textbf{0.765} & \textbf{14.436} & \textbf{0.861} & \textbf{0.786} & \textbf{0.732} & \textbf{0.513} & \textbf{0.96} \\
\bottomrule
\end{tabular}
\caption{Component Ablation of the MolGVR Pipeline.
Pass@1 and pass@5 indicate that the Generator produces one or five candidates for each input.
For the full pipeline, each generated candidate is checked by the Verifier; rejected candidates are refined once by the Refiner, and then merged with the initially accepted ones for final evaluation.
The full pipeline consistently improves Match under both pass@1 and pass@5, while also improving most similarity metrics.
Best results are highlighted in bold and second-best results are underlined.}
\label{tab:ablation}
\end{table*}

\begin{table*}
    \centering
    \small
    \setlength{\tabcolsep}{2pt}
    \begin{tabular}{llccccccccc}
        \toprule
        Dataset
        & Method
        & Validity$\uparrow$
        & BLEU$\uparrow$
        & Levenshtein$\downarrow$
        & MACCS FTS$\uparrow$
        & RDK FTS$\uparrow$
        & Morgan FTS$\uparrow$
        & Match$\uparrow$
        & FCD$\downarrow$
        & Violation$\downarrow$ \\
        \midrule

        \multirow{4}{*}{ChEBI-20}
        & Chem-R-8B (pass@1)
        & 0.935 & 0.824 & 19.830 & 0.917 & 0.830 & 0.777 & 0.414 & 8.41 & 8.81\% \\

        & + V\&R
        & \underline{0.979} & 0.850 & 17.089 & 0.914 & 0.824 & 0.768 & 0.419 & \textbf{4.06} & \underline{3.81\%} \\

        & + MSR + RL
        & 0.967 & \underline{0.861} & \underline{14.072} & \textbf{0.945} & \textbf{0.889} & \textbf{0.845} & \underline{0.583} & \underline{5.27} & 5.03\% \\

        & + MSR + RL + V\&R
        & \textbf{0.988} & \textbf{0.872} & \textbf{13.303} & \underline{0.942} & \underline{0.884} & \underline{0.839} & \textbf{0.586} & 5.32 & \textbf{2.24\%} \\

        \midrule

        \multirow{4}{*}{PCDes}
        & Chem-R-8B (pass@1)
        & 0.941 & 0.649 & 24.721 & 0.808 & 0.716 & 0.654 & 0.411 & \underline{0.74} & 5.73\% \\

        & + V\&R
        & \underline{0.985} & 0.703 & 20.936 & 0.802 & 0.707 & 0.642 & 0.414 & \textbf{0.69} & \underline{1.40\%} \\

        & + MSR + RL
        & 0.976 & \underline{0.717} & \underline{17.110} & \textbf{0.842} & \textbf{0.767} & \textbf{0.711} & \underline{0.502} & 3.76 & 2.36\% \\

        & + MSR + RL + V\&R
        & \textbf{0.994} & \textbf{0.723} & \textbf{16.463} & \underline{0.840} & \underline{0.765} & \underline{0.707} & \textbf{0.503} & 3.81 & \textbf{0.50\%} \\

        \bottomrule
    \end{tabular}
    \caption{
    Generalization across Other Models using Chem-R-8B as backbones. ``MSR + RL'' denotes
    the generator trained with our proposed strategy.
    ``V\&R'' denotes applying the Verifier and Refiner.
    Violation denotes the percentage of test samples containing verifier-detected violation.
    Best results are highlighted in bold, and second-best results are underlined.
    }
    \label{tab:cross_backbone_violation_analysis}
\end{table*}

The improvements are also evident on PCDes.
MolGVR achieves the best Match score of 0.500, surpassing bioT5 (0.455) and MolT5$_{\text{large}}$ (0.440).
It further obtains the best MACCS, RDK, and Morgan fingerprint similarities among the compared methods. 

On PCDes, MolGVR also shows remaining gaps on validity, BLEU, Levenshtein distance, and FCD.
These gaps are consistent with the observations on ChEBI-20: validity is already strong, BLEU and Levenshtein distance are sensitive to the SMILES strings, and FCD reflects distribution-level alignment rather than instance-level molecular correctness.

A PCDes example is shown in Figure~\ref{fig:demo}(b). The Verifier extracts a group-count constraint requiring two hydroxy groups, and a monocarboxylic-acid constraint. The initial candidate lacks the required side-chain hydroxyl group and therefore violates the dihydroxy constraint. Using this failure feedback, the Refiner restores the missing hydroxyl group and produces the correct molecule. 
Additional examples are provided in Appendix~\ref{sec:more_exp_pcdes}.

\subsection{Ablation Study}

\subsubsection{Component Ablation of the MolGVR Pipeline}

To examine the contributions of different components in MolGVR, Table~\ref{tab:ablation} compares the standalone Generator, the Generator followed by indiscriminate Refiner, and the full verification-guided pipeline. The standalone Generator is already strong after MSR SFT\&RL, and a detailed ablation of MSR SFT\&RL is provided in Appendix Table~\ref{tab:generator_training_ablation}. However, directly applying the Refiner to all generated candidate substantially degrades performance. This indicates that Refiner may alter candidates that already satisfy the description. In contrast, the full pipeline uses the Verifier to selectively refine only rejected candidates, improving Match and most validity and similarity metrics. These results demonstrate that the gains arise from verification-guided, targeted refinement rather than refinement alone.

\subsubsection{Verification-Guided Refinement vs.\ Additional Sampling}

\begin{figure}[t]
\centering
\includegraphics[width=0.9\columnwidth]
{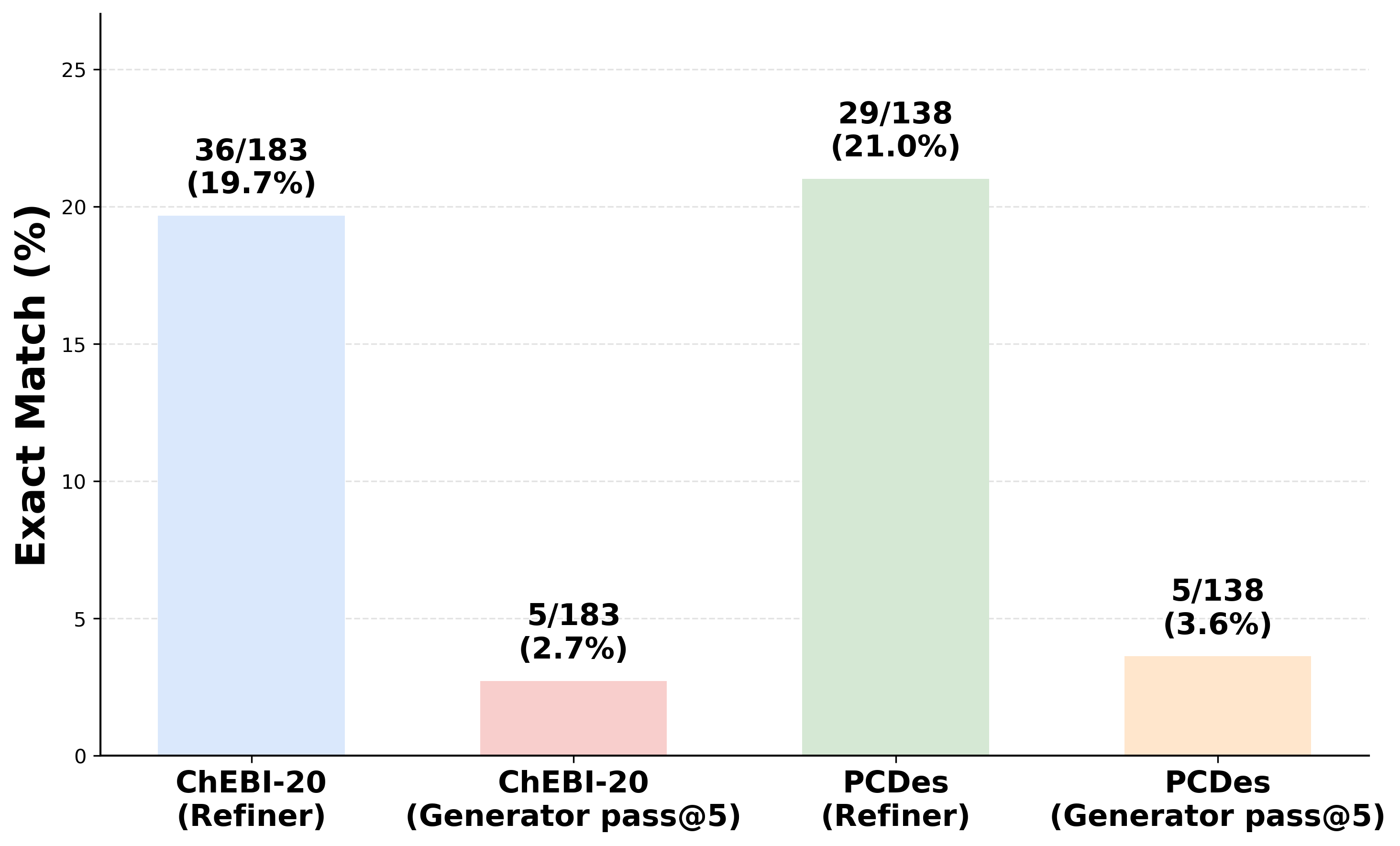}

\caption{Correction rates of verification-guided refinement and five additional Generator samples on verifier-rejected candidates.}
\label{fig:refiner_ablation}
\end{figure}

We further examine whether the improvement of the full pipeline comes merely from additional generation.
Under the pass@1 setting, the Verifier identifies 183/1414 rule-violating candidates on ChEBI-20 and 138/1526 on PCDes. A detailed analysis of the Verifier's decision behavior is provided in Appendix~\ref{sec:app_eval_verifier}.
On these rejected candidates, Figure~\ref{fig:refiner_ablation} compares verification-guided correction with continued sampling from the Generator under a pass@5 setting.
On ChEBI-20, the Refiner corrects 36 out of 183 rejected candidates (19.7\%), whereas additional Generator sampling recovers only 5 out of 183 cases (2.7\%).
On PCDes, the Refiner corrects 29 out of 138 rejected candidates (21.0\%), while the Generator recovers only 5 out of 138 cases (3.6\%).

These results show that targeted correction addresses verifier-detected failure modes more effectively than additional sampling, thereby mitigating the verification-and-correction gap.
Moreover, the additional gains under pass@5 indicate that refinement and multi-sample generation are complementary: sampling increases candidate diversity, while refinement corrects verifier-identified failures.
We additionally report the inference time of MolGVR under pass@1 and pass@5 in ~\ref{sec:app_inference_time}.

\subsubsection{Generalization across Other Models}

Table~\ref{tab:cross_backbone_violation_analysis} evaluates our method on Chem-R-8B, an chemistry foundation model backbone. MSR+RL consistently improves generation quality, increasing Match from 0.414 to 0.583 on ChEBI-20 and from 0.411 to 0.502 on PCDes. Applying V\&R to either the original or trained generator further reduces violation rates while improving validity and Match. Consequently, the full pipeline achieves the best validity, BLEU, Levenshtein, Match, and violation rate on both datasets. These results demonstrate that MSR+RL and V\&R provide complementary gains across generator backbones: MSR+RL improves the generator itself, while V\&R further enhances performance by reducing the proportion of invalid or structurally noncompliant outputs.

\subsubsection{Category-wise Analysis of Verification and Refinement}

\begin{table}[h]
\centering
\small
\setlength{\tabcolsep}{1pt}
\begin{tabular}{llrrr}
\toprule
Dataset & Constraint Type & G. Vio$\downarrow$ & +V\&R. Vio$\downarrow$ & Repair Rate$\uparrow$ \\
\midrule
\multirow{7}{*}{ChEBI-20}
& Functional group & 35 & 10 & 71.4 \\
& Core scaffold\&ring & 15 & 9 & 40.0 \\
& Element\&atom type & 5 & 2 & 60.0 \\
& Element count & 13 & 9 & 30.8 \\
& Group count & 15 & 7 & 53.3 \\
& Stereochemistry & 30 & 11 & 63.3 \\
& Invalid SMILES & 75 & 25 & 66.7 \\
\midrule
\multirow{7}{*}{PCDes}
& Functional group & 44 & 16 & 63.6 \\
& Core scaffold\&ring & 14 & 6 & 57.1 \\
& Element\&atom type & 5 & 5 & 0.0 \\
& Element count & 11 & 7 & 36.4 \\
& Group count & 7 & 2 & 71.4 \\
& Salt\&charge state & 17 & 14 & 17.6 \\
& Invalid SMILES & 48 & 10 & 79.2 \\
\bottomrule
\end{tabular}
\caption{Category-wise breakdown of verifier-detected violations and the corresponding refinement outcomes. Repair Rate measures the proportion of verifier-detected noncompliance cases corrected by the Refiner.}
\label{tab:constraint_breakdown}
\end{table}

To assess MolGVR across error types, Table~\ref{tab:constraint_breakdown} reports the violations detected by the Verifier and the corresponding refinement outcomes. The Refiner removes 61.2\% and 58.9\% of detected violations on ChEBI-20 and PCDes, respectively. It performs particularly well on functional-group constraints and invalid SMILES, achieving Repair Rates of 63.6--71.4\% and 66.7--79.2\%, respectively. Strong results are also observed for stereochemistry on ChEBI-20 and group-count constraints on PCDes. Although exact-molecule correction is more stringent, MolGVR still fixes approximately 20\% of rejected candidates on both datasets. Core-scaffold and salt/charge-state errors remain more difficult, as they often require broader structural modifications. These results show that MolGVR improves generation by detecting and correcting outputs that fail verification.

\section{Conclusion}

We presented MolGVR, a chemistry-grounded Generator--Verifier--Refiner framework for text-to-molecule generation. By moving beyond one-shot generation, MolGVR combines an evidence-driven Generator, an LLM-assisted and RDKit-based Verifier, and a Refiner that revises incorrect molecules based on failure feedback provided by the Verifier. Experiments on ChEBI-20 and PCDes show that MolGVR improves exact match and related similarity metrics. Qualitative examples and ablation studies show that these gains arise from effective verification and feedback-guided correction rather than repeated sampling alone. Our results suggest that explicit chemical verification and refinement mitigate the verification-and-correction gap and provide a promising direction for improving structural fidelity in text-to-molecule generation.

\bibliography{custom}

\newpage
\appendix
\setcounter{secnumdepth}{2}
\section{More details about MolGVR Generator}

\subsection{Reward Design}
\label{sec:app_reward_design}
Our RL reward is composed of a \emph{format reward} and an \emph{accuracy reward}. The overall reward is defined as
\begin{equation}
r(o)
=
(1-\lambda)\,r_{\mathrm{acc}}(o)
+
\lambda\,r_{\mathrm{fmt}}(o),
\end{equation}
where $\lambda=0.1$ in our implementation.

\paragraph{Format reward.}

Let $\mathcal{F}_{\mathrm{full}}$ denote the set of responses that contain all six evidence tags in the prescribed order, followed by \texttt{</think>} and a final boxed answer, and let $\mathcal{F}_{\mathrm{partial}}$ denote the set of responses that contain \texttt{</think>} and a final boxed answer but do not satisfy the full tagged format. We define the format reward as
\begin{equation}
r_{\mathrm{fmt}}(o)=
\begin{cases}
1, & \text{if } o \in \mathcal{F}_{\mathrm{full}},\\
0.2, & \text{if } o \in \mathcal{F}_{\mathrm{partial}},\\
0, & \text{otherwise.}
\end{cases}
\end{equation}
This shaped reward encourages the model to preserve a parsable reasoning structure even when the final answer is incorrect.

\paragraph{Accuracy reward.}
The accuracy reward is a binary exact-match reward over the extracted final boxed answer:
\begin{equation}
r_{\mathrm{acc}}(o)
=
\begin{cases}
1.0, & \text{if } \mathrm{ans}(o) = \mathrm{ans}(y^\star),\\
0.0, & \text{otherwise,}
\end{cases}
\end{equation}
where $\mathrm{ans}(\cdot)$ extracts the content inside the final \texttt{\textbackslash boxed\{\}} span. Our RL stage optimizes a string-level exact-match reward between the extracted prediction and the gold answer. This raw-string reward is used for computational efficiency, since GRPO requires evaluating rewards over many sampled rollouts and repeated RDKit-based canonicalization would be costly.

\subsection{Ablation of Generator Training Components}

\begin{table*}[t]
\centering
\scriptsize
\resizebox{\textwidth}{!}{
\begin{tabular}{llcccccccccc}
\toprule
Dataset & Setting
& MSR & RL
& Validity$\uparrow$
& BLEU$\uparrow$
& Lev.$\downarrow$
& MACCS$\uparrow$
& RDK$\uparrow$
& Morgan$\uparrow$
& Match$\uparrow$
& FCD$\downarrow$ \\
\midrule
\multirow{8}{*}{ChEBI-20}
& Raw Generator (pass@1)
& \xmark & \xmark
& 0.966 & 0.832 & 17.952 & 0.922 & 0.842 & 0.787 & 0.448 & 2.54 \\

& + RL
& \xmark & \cmark
& 0.972 & 0.605 & 47.673 & 0.938 & 0.873 & 0.829 & 0.547 & 56.78 \\

& + MSR
& \cmark & \xmark
& 0.958 & 0.854 & 16.369 & 0.931 & 0.851 & 0.802 & 0.464 & 3.05 \\

& + MSR + RL
& \cmark & \cmark
& 0.977 & 0.877 & 13.010 & \underline{0.947} & \underline{0.887} & 0.844 & \underline{0.571} & 3.94 \\

& Raw Generator (pass@5)
& \xmark & \xmark
& \textbf{0.993} & 0.878 & 13.310 & 0.941 & 0.878 & 0.831 & 0.524 & \textbf{1.45} \\

& + RL
& \xmark & \cmark
& 0.984 & 0.876 & 13.165 & 0.943 & 0.884 & 0.841 & 0.567 & \underline{1.67} \\

& + MSR
& \cmark & \xmark
& \underline{0.992} & \underline{0.891} & \underline{12.001} & 0.946 & 0.885 & \underline{0.845} & 0.552 & 2.19 \\

& + MSR + RL
& \cmark & \cmark
& 0.991 & \textbf{0.893} & \textbf{11.563} & \textbf{0.952} & \textbf{0.898} & \textbf{0.856} & \textbf{0.593} & 2.44 \\

\midrule
\multirow{8}{*}{PCDes}
& Raw Generator (pass@1)
& \xmark & \xmark
& 0.974 & 0.687 & 20.623 & 0.819 & 0.721 & 0.656 & 0.385 & 1.24 \\

& + RL
& \xmark & \cmark
& 0.982 & 0.709 & 17.862 & 0.835 & 0.757 & 0.699 & 0.474 & 1.13 \\

& + MSR
& \cmark & \xmark
& 0.967 & 0.698 & 19.367 & 0.824 & 0.721 & 0.656 & 0.392 & \textbf{0.77} \\

& + MSR + RL
& \cmark & \cmark
& 0.985 & 0.733 & 16.538 & 0.845 & 0.767 & 0.708 & \underline{0.490} & \underline{0.97} \\

& Raw Generator (pass@5)
& \xmark & \xmark
& \underline{0.996} & 0.736 & 16.463 & 0.850 & 0.767 & 0.711 & 0.450 & 1.11 \\

& + RL
& \xmark & \cmark
& 0.993 & 0.733 & 16.350 & 0.846 & \underline{0.769} & \underline{0.714} & 0.488 & 1.09 \\

& + MSR
& \cmark & \xmark
& \textbf{0.997} & \underline{0.749} & \underline{16.045} & \underline{0.853} & 0.765 & 0.711 & 0.452 & 1.03 \\

& + MSR + RL
& \cmark & \cmark
& \underline{0.996} & \textbf{0.756} & \textbf{15.104} & \textbf{0.858} & \textbf{0.781} & \textbf{0.725} & \textbf{0.503} & \underline{0.97} \\

\bottomrule
\end{tabular}
}
\caption{
Ablation of MSR and RL for standalone Generator training under pass@1 and pass@5.
All rows report Generator-only results without applying the Verifier or Refiner.
MSR denotes supervised training with intermediate molecular structural reasoning targets, while RL denotes GRPO training with accuracy and format rewards.
Combining MSR and RL achieves the strongest overall Generator performance on both datasets.
Best results are highlighted in bold and second-best results are underlined.
}
\label{tab:generator_training_ablation}
\end{table*}

Table~\ref{tab:generator_training_ablation} evaluates the individual and combined effects of MSR and RL on Generator training. Both components generally improve molecular generation, while their combination achieves the strongest overall performance under both pass@1 and pass@5 on the two datasets. In particular, MSR+RL improves Match from 0.448 to 0.571 under pass@1 and from 0.524 to 0.593 under pass@5 on ChEBI-20. Similar improvements are observed on PCDes, where Match increases from 0.385 to 0.490 and from 0.450 to 0.503, respectively. The combined setting also achieves the best results on most string-similarity and fingerprint-similarity metrics, string-similarity, and fingerprint-similarity metrics, demonstrating that MSR supervision and RL provide complementary benefits for Generator training.

\section{More details about MolGVR Verifier}
\subsection{Rule Extraction}
\label{sec:app_verify_rule}

\paragraph{Dataset-specific verification schemas.}
The Verifier supports seven noncompliance categories for each dataset. Six categories are derived from molecular descriptions, while invalid SMILES is included as a built-in category determined directly through RDKit parsing.

For ChEBI-20, the verification categories are:
\begin{multline}
R_{\text{ChEBI}} =
\{
R_{\text{fg}},
R_{\text{scaffold}},
R_{\text{elem}}, \\
R_{\text{elem-count}},
R_{\text{group-count}},
R_{\text{stereo}},
R_{\text{invalid}}
\},
\end{multline}
corresponding to \textit{functional groups}, \textit{core scaffolds or ring systems}, \textit{elements or atom types}, \textit{element count requirements}, \textit{group count requirements}, \textit{stereochemistry requirements}, and \textit{invalid SMILES}.

For PCDes, the verification categories are:
\begin{multline}
R_{\text{PCDes}} =
\{
R_{\text{fg}},
R_{\text{scaffold}},
R_{\text{elem}}, \\
R_{\text{elem-count}},
R_{\text{group-count}},
R_{\text{salt/charge}},
R_{\text{invalid}}
\},
\end{multline}
where \textit{salt or charge state} replaces stereochemistry. The first six categories are extracted from the molecular description. The invalid-SMILES category does not require extraction and is triggered when a candidate cannot be parsed by RDKit.

\paragraph{Conservative normalization.}
The extractor is only allowed to output normalized labels from a closed vocabulary. This reduces variation in surface forms and makes downstream verification executable. At the same time, the prompt explicitly discourages over-normalization: if a phrase cannot be mapped reliably to one of the allowed normalized labels, it is omitted. This conservative design is important for two reasons. First, the verifier is intended to produce \emph{actionable} constraints rather than broad semantic summaries. Second, a more permissive extractor would increase false-positive verification failures by introducing unverifiable or weakly grounded rules.

\paragraph{Rule Extraction Prompt.}\mbox{}\\
The prompts below extract only the six description-derived categories for each dataset. Invalid SMILES is handled separately as a built-in seventh category during candidate parsing and therefore does not appear in the extraction schema.

\noindent\textbf{ChEBI-20:}
\begin{myverb}
You are an information extraction model for molecule-description verification.

Your task is to extract ONLY explicit, directly stated, structure-related information about the TARGET molecule from a molecule description.

Be conservative:
- Do NOT infer.
- Do NOT guess.
- Do NOT use chemistry background knowledge to fill in missing facts.
- Do NOT convert biological roles, uses, sources, properties, salts, acid/base context, or references to other molecules into structural constraints.
- Extract ONLY properties that are explicitly asserted about the target molecule itself.
- If something is not explicitly stated about the target molecule, do not extract it.
- When uncertain, omit rather than guess.

Target molecule only:
- The description may mention source molecules, parent molecules, precursor molecules, constituent molecules, residues, moieties, substituent donors, metabolites, conjugate acids/bases, salts, hydrates, or compounds from which the target molecule is derived.
- Do NOT transfer any property from those referenced molecules/components onto the target molecule unless the text explicitly states that the target molecule itself has that property.
- In particular, do NOT treat descriptions of source or component molecules as descriptions of the whole target molecule.
- If the text says the molecule is derived from, formed from, resulting from condensation of, esterified from, amidated from, conjugated to, substituted by, metabolite of, salt of, hydrate of, or related to another molecule, do NOT extract structural properties that belong only to that other molecule unless they are explicitly re-stated as properties of the target molecule.
- You may extract a feature only if the target molecule itself is explicitly named as having that feature in the allowed categories.
- Example: if the text says "derived from dodecanoic acid", do NOT extract carboxylic_acid for the target molecule from that phrase alone.
- Example: if the text says "an ester resulting from condensation of two molecules of dodecanoic acid", you may extract ester if explicitly stated, but do NOT transfer "carboxylic acid" from dodecanoic acid to the target molecule.
- Example: if the text says "X substituted by a quinoline group", do NOT automatically extract quinoline unless the target molecule itself is explicitly described using an allowed scaffold term in a way that directly applies to the whole structure and is reliably checkable.

Extract ONLY information that can be checked reliably using RDKit or simple rule-based matching.

Extract these categories:

1. functional_groups
   Allowed normalized values:
   ester, amide, carboxylic_acid, carboxylate, ketone, aldehyde, alcohol,
   phenol, amine, sulfonamide, carbamate, phosphate, phosphonate, sulfate,
   nitrile, ether, thiol, sulfone, thioester, hydroxamate, imide, amidine,
   amidinium, carboxamidinium, sulfonic_acid, sulfonate, phosphodiester,
   phosphocholine, phenoxide, alkoxide, gem_diol, alpha_halocarbonyl

2. core_scaffolds_or_ring_systems
   Allowed normalized values:
   benzene, pyridine, quinoline, isoquinoline, triazole, tetrazole,
   piperidine, benzofuran, chromene, pyrimidine, oxazolidinone,
   phenothiazine, indole, naphthalene, purine, uracil, benzoxazole,
   benzothiazole, oxazole, thiazole, thiazolium, imidazole, steroid, pregnane

3. elements_or_atom_types
   Allowed normalized values:
   fluorine, chlorine, bromine, iodine, phosphorus, sulfur, selenium,
   sodium, potassium, calcium, iron, cobalt, zinc, platinum, arsenic,
   nitrogen, oxygen

4. element_count_requirements
   Extract only explicit counts of elements or atom types.
   Allowed normalized values:
   fluorine, chlorine, bromine, iodine, phosphorus, sulfur, selenium,
   sodium, potassium, calcium, iron, cobalt, zinc, platinum, arsenic,
   nitrogen, oxygen
   Examples:
   one phosphorus atom, two nitrogen atoms, three chlorine atoms, dichloro

5. group_count_requirements
   Extract only explicit counts of functional groups or directly checkable subunits.
   Allowed normalized values:
   oxo_group, hydroxy_group, amino_group, carboxylic_acid, carboxylate,
   amide, ester, thioester, phosphate, phosphonate, sulfate, sulfonamide,
   nitrile, ether, double_bond, c_c_double_bond
   Examples:
   two oxo groups, one double bond, two hydroxy groups
   Do NOT extract counts that require uncertain residue decomposition, such as
   trisaccharide, tetrasaccharide, polymer length, PEG length, or other complex subunit counting.

6. stereochemistry_requirements
   Extract ONLY stereochemical information that can be checked without parent-structure numbering.
   Allowed normalized values:
   r_center, s_center, e_double_bond, z_double_bond, cis_double_bond, trans_double_bond
   Allowed examples:
   one R center, two S centers, one E double bond
   Do NOT extract:
   - position-dependent stereochemistry, such as:
     "S at position 1", "R at C-4"
   - molecule-level labels that are not reliably verifiable from a single SMILES, such as:
     racemate, meso, enantiomer
   - alpha/beta descriptors that require attachment/numbering context

Return valid JSON only, with exactly this schema:

{{
  "functional_groups": [
    {{
      "text_span": "",
      "normalized": ""
    }}
  ],
  "core_scaffolds_or_ring_systems": [
    {{
      "text_span": "",
      "normalized": ""
    }}
  ],
  "elements_or_atom_types": [
    {{
      "text_span": "",
      "normalized": ""
    }}
  ],
  "element_count_requirements": [
    {{
      "text_span": "",
      "normalized": "",
      "count": 0
    }}
  ],
  "group_count_requirements": [
    {{
      "text_span": "",
      "normalized": "",
      "count": 0
    }}
  ],
  "stereochemistry_requirements": [
    {{
      "text_span": "",
      "normalized": "",
      "count": 0
    }}
  ]
}}

Rules:
- "text_span" must be a short exact phrase copied from the description.
- "normalized" must be chosen ONLY from the allowed values listed above for that category.
- Do NOT invent new normalized labels.
- If no allowed normalized value fits exactly, omit the item.
- Do not over-normalize.
- If a category is not mentioned, return an empty list for that category.
- For element_count_requirements, group_count_requirements, and stereochemistry_requirements, use a positive integer in "count".
- If the count is not explicit, do not extract it.
- If a phrase refers to another molecule, precursor, source, residue, substituent donor, parent compound, or component rather than the target molecule itself, do not extract it.

Do NOT extract:
- biological role
- pharmacological activity
- therapeutic use
- source organism
- biomarker claims
- odor, taste, solubility, melting point, flash point
- location or clinical statements
- salts, hydrates, acid/base context, protonation state
- conjugate acid / conjugate base
- major species at pH
- family/class labels
- parent-compound context
- source-molecule descriptors
- component-molecule descriptors
- precursor or constituent descriptors
- residue/moiety descriptions unless explicitly asserted as an allowed, target-molecule-level feature
- "used as ..."
- "derived from ..."
- "formed from ..."
- "resulting from condensation of ..."
- "ester of ..."
- "amide of ..."
- "salt of ..."
- "hydrate of ..."
- "metabolite of ..."
- "replaced by ..."
- position-dependent attachment or numbering information
- position-dependent stereochemistry

If nothing clearly belongs to these categories, return all lists as empty lists.

Description: {des}
\end{myverb}

\noindent\textbf{PCDes:}
\begin{myverb}
You are an information extraction model for molecule-description verification on the PCDes dataset.

Your task is to extract ONLY explicit, directly stated, structure-relevant information about the TARGET molecule that can be USED DIRECTLY in downstream verification.

Core rule:
- Extract a fact ONLY if it is:
  1) explicitly stated about the target molecule, and
  2) directly verifiable from the molecule structure / charge / disconnected components.
- If a fact would likely be discarded later because it is too ambiguous, too ontology-like, too position-dependent, too context-dependent, or too hard to verify reliably, DO NOT extract it.
- When uncertain, omit rather than guess.

Target molecule only:
- The description may mention source molecules, parent molecules, precursor molecules, conjugate acids/bases, metabolites, salts, or compounds from which the target molecule is derived.
- Extract a feature only if the TARGET molecule itself is explicitly described as having that feature.
- Do NOT transfer structural features from parent/source/reference molecules to the target molecule.
- Example: "derived from alanine" does NOT justify extracting alanine-like structure.
- Example: "hydrochloride salt form of X" DOES justify extracting hydrochloride_salt because that directly describes the target molecule.
- Example: "it is a member of triazoles" DOES justify extracting triazole because that directly describes the target molecule.

Do NOT infer:
- Do NOT use chemistry background knowledge to fill in missing facts.
- Do NOT infer hidden substructures from broad biological or ontology language.
- Do NOT infer charge state from pKa, "acidic/basic", or physiological statements unless the description explicitly states an anion/cation/zwitterion/salt form.
- Do NOT infer position-specific substituents into general rules unless the resulting feature itself is explicitly named in an allowed category.

Extract ONLY the following categories.

1. functional_groups
Allowed normalized values:
carboxylic_acid, carboxylate, amide, ester, carbamate, ketone, aldehyde,
alcohol, phenol, amine, ether, nitrile, nitroso, sulfonic_acid,
sulfonamide, phosphate, phosphodiester, isourea

Extract only if the functional group is directly and reliably checkable as a concrete substructure.

2. core_scaffolds_or_ring_systems
Allowed normalized values:
benzene, pyridine, pyrimidine, imidazole, triazole, quinoline,
benzodiazepinone, naphthalene, steroid

Extract only if the scaffold/ring system is explicitly stated for the target molecule.
Do NOT map an unknown class name to the closest allowed label.

3. elements_or_atom_types
Allowed normalized values:
fluorine, chlorine, iodine, phosphorus, sulfur, sodium, calcium,
strontium, mercury, molybdenum, nitrogen, oxygen

Extract only if the target molecule is explicitly described as containing that element or atom type.

4. element_count_requirements
Allowed normalized values:
fluorine, chlorine, iodine, phosphorus, sulfur, sodium, calcium,
strontium, mercury, molybdenum, nitrogen, oxygen

Extract only if an explicit count is directly stated and directly usable.
Examples:
- one phosphorus atom
- two chlorine atoms
- 1:1 ratio of calcium and oxygen

If the count is not explicit, do not extract it.

5. group_count_requirements
Allowed normalized values:
carboxylic_acid, carboxylate, hydroxy_group

Extract only if the count is explicit and directly usable.
Examples:
- monocarboxylic acid -> carboxylic_acid count = 1
- dicarboxylic acid -> carboxylic_acid count = 2
- tricarboxylic acid -> carboxylic_acid count = 3
- dihydroxy -> hydroxy_group count = 2
- monohydroxy -> hydroxy_group count = 1

Do NOT infer counts from class names unless the count is explicit in the wording.

6. salt_or_charge_state
Allowed normalized values:
hydrochloride_salt, mesylate_salt, sodium_salt,
zwitterion, monoanion, dianion, trianion

Extract only if the target molecule is explicitly described in that form.
Examples:
- hydrochloride salt form
- mesylate salt form
- sodium salt
- zwitterionic form
- monocarboxylic acid anion
- dicarboxylate anion
- trianion

Do NOT extract:
- vague acid/base language
- conjugate acid/conjugate base statements unless the target molecule itself is explicitly described using one of the allowed charge-state labels above

Return valid JSON only, with exactly this schema:

{{
  "functional_groups": [
    {{
      "text_span": "",
      "normalized": ""
    }}
  ],
  "core_scaffolds_or_ring_systems": [
    {{
      "text_span": "",
      "normalized": ""
    }}
  ],
  "elements_or_atom_types": [
    {{
      "text_span": "",
      "normalized": ""
    }}
  ],
  "element_count_requirements": [
    {{
      "text_span": "",
      "normalized": "",
      "count": 0
    }}
  ],
  "group_count_requirements": [
    {{
      "text_span": "",
      "normalized": "",
      "count": 0
    }}
  ],
  "salt_or_charge_state": [
    {{
      "text_span": "",
      "normalized": ""
    }}
  ]
}}

Global rules:
- "text_span" must be a short exact phrase copied verbatim from the description.
- "normalized" must be chosen ONLY from the allowed values for that category.
- Do NOT invent new normalized labels.
- If no allowed label fits exactly, omit the item.
- Do NOT over-normalize.
- If a category is absent, return an empty list for that category.
- For count fields, "count" must be a positive integer.
- If the count is not explicit, do not extract it.
- If the same fact appears multiple times, keep only one copy.
- If the phrase refers to another molecule rather than the target molecule, do not extract it.

Do NOT extract:
- biological role
- pharmacological activity
- therapeutic use
- assay mechanism
- source organism
- biomarker claims
- subcellular location
- odor, taste, solubility, melting point, boiling point, density, flash point
- clinical or formulation-only statements that do not encode a concrete structural fact
- "used as ..."
- "has a role as ..."
- "detected in ..."
- "located in ..."
- "derived from ..."
- "metabolite of ..."
- "analogue of ..."
- "related to ..."
- parent/source/precursor context
- family labels not present in the allowed lists
- residue decomposition
- sugar-unit identity
- chain length
- position-specific substitution statements
- position-numbered attachment information
- ALL stereochemistry information, including:
  - R/S labels
  - E/Z labels
  - cis/trans labels
  - alpha/beta descriptors
  - enantiomer / diastereomer / optically active
  - anomeric configuration

If nothing clearly belongs to the allowed directly usable categories, return all lists as empty lists.

Description: {des}

\end{myverb}

\subsection{Tool-based Verification}
\label{sec:app_verify_tool}

\paragraph{Rule categories and checker design.}
Across the two datasets, the Verifier supports eight distinct noncompliance categories. Each dataset uses seven categories: six shared categories and one dataset-specific category, namely stereochemistry for ChEBI-20 or salt and charge state for PCDes.

\begin{itemize}
    \item \textbf{Invalid SMILES.}
    Each candidate is first parsed using \texttt{Chem.MolFromSmiles}. A candidate is assigned to the invalid-SMILES category when parsing fails. This built-in check is shared by ChEBI-20 and PCDes and does not require a description-derived rule.

    \item \textbf{Functional groups.}
    Functional groups are checked by SMARTS-based substructure matching, including carboxylic acids, carboxylates, amides, esters, aldehydes, ketones, nitriles, sulfonamides, phosphates, phosphodiesters, and others. For broader categories such as alcohol, amine, or ether, we additionally use custom counting heuristics.

    \item \textbf{Core scaffolds and ring systems.}
    Ring systems such as benzene, pyridine, pyrimidine, imidazole, triazole, quinoline, and naphthalene are verified by SMARTS patterns. For more complex structural families, such as steroid-like fused ring systems or benzodiazepinone-like cores, we use dedicated topology-based heuristics instead of relying on a single SMARTS rule.

    \item \textbf{Element or atom type.}
    Element-presence constraints are verified by mapping each normalized element label to its atomic symbol and counting atoms with that symbol in the RDKit molecular graph. A presence requirement is satisfied when at least one corresponding atom is found.

    \item \textbf{Element count.}
    Explicit element-count constraints are verified using the same atom-symbol counts. The number of atoms whose symbols match the normalized element must equal the extracted \texttt{count}; only expert-validated count checks are allowed to trigger rejection.

    \item \textbf{Group counts.}
    Explicit counts of groups such as hydroxy groups, amino groups, carboxylates, oxo groups, or carbon--carbon double bonds are verified using RDKit-based counting functions tailored to each group type.

    \item \textbf{Stereochemistry (ChEBI-20).}
    For ChEBI-20, stereochemical constraints such as the number of \texttt{R} or \texttt{S} centers and the number of \texttt{E}/\texttt{Z} double bonds are verified after RDKit stereochemistry assignment.

    \item \textbf{Salt and charge state (PCDes).}
    For PCDes, rules such as \texttt{hydrochloride\_salt}, \texttt{sodium\_salt}, \texttt{mesylate\_salt}, \texttt{monoanion}, \texttt{dianion}, or \texttt{zwitterion} are verified by combining RDKit-based disconnected-component analysis with atom- and component-level formal-charge checks.
\end{itemize}

\section{Experimental Setup}
\subsection{Dataset and Split}
\label{sec:app_data}

We conduct experiments on two text-to-molecule generation benchmarks, ChEBI-20~\cite{edwards2022translation} and PCDes~\cite{zeng2022deep}.
For ChEBI-20, we follow the standard train/validation/test split used in prior text-to-molecule generation studies.
For PCDes, we follow the preprocessed split adopted by LDMol~\cite{chang2025ldmol}, resulting in 8,983 training pairs and 2,998 test pairs.
Models are trained on the corresponding training split and evaluated on the test split of each dataset.

\begin{table}[t]
\centering
\small
\begin{tabular}{lrrrr}
\toprule
Dataset & Train & Validation & Test & Total \\
\midrule
ChEBI-20 & 26,407 & 3,301 & 3,300 & 33,008 \\
PCDes    & 8,983  & -- & 2,998 & 11,981 \\
\bottomrule
\end{tabular}
\caption{Statistics of the datasets used in our experiments. For PCDes, we follow the LDMol-preprocessed split.}
\label{tab:dataset_statistics}
\end{table}

\subsection{Data Synthesis Details}
\label{sec:app_data_synthesis}
\begin{figure*}[!t]
\centering
\includegraphics[width=\textwidth]{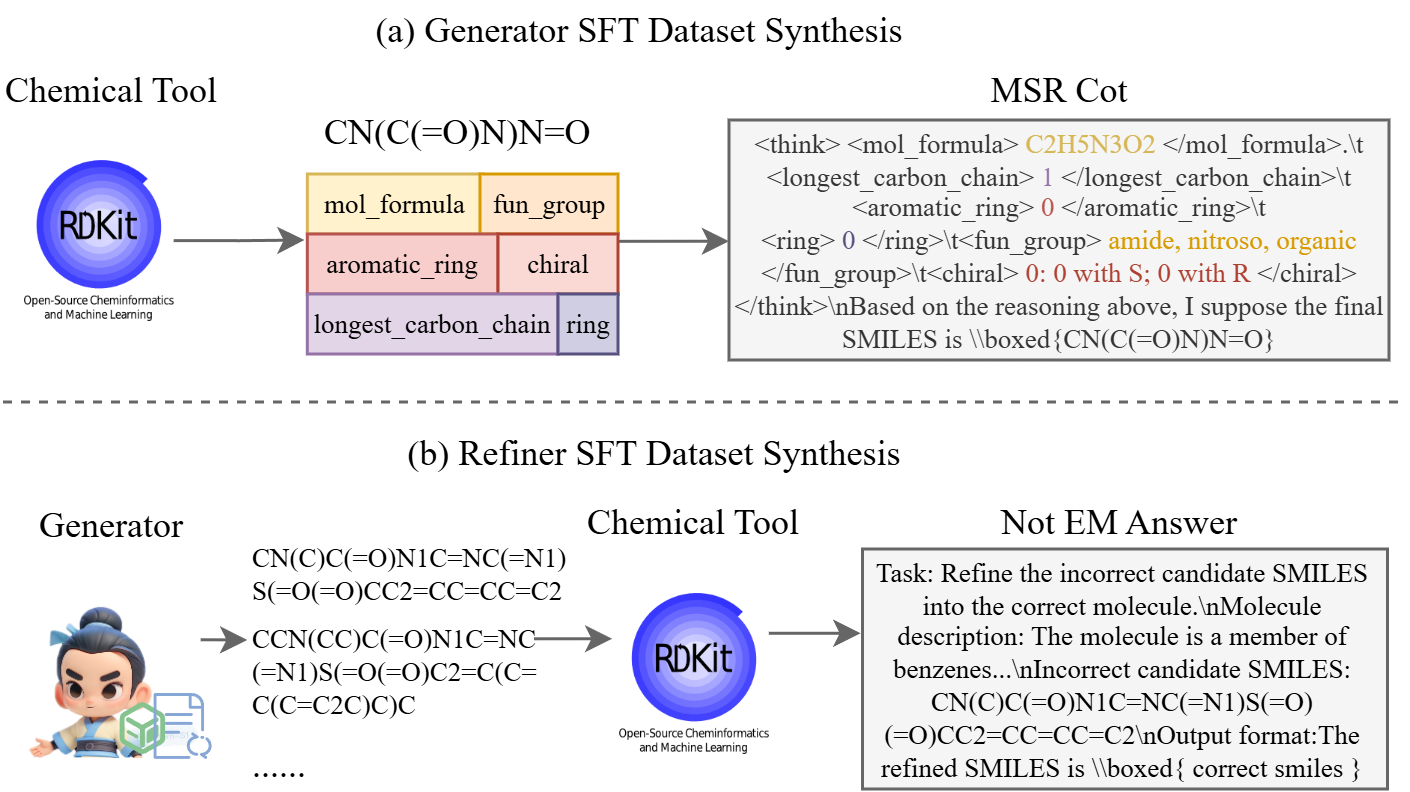}

\caption{Data synthesis process.
(a) Generator SFT data are constructed by using RDKit to extract six types of structural evidence from the gold SMILES and converting them into a tagged MSR-style chain of thought followed by the correct molecular answer.
(b) The first-stage Refiner SFT data are constructed from generator-induced errors: sampled candidates that do not exactly match the gold molecule are paired with the original description, while the correct molecule serves as the supervised correction target.}
\label{fig:data}
\end{figure*}

\paragraph{Generator SFT data synthesis.}
As shown in Figure~\ref{fig:data}(a), we construct the Generator SFT targets by augmenting each gold molecule with explicit structural evidence. For each training example, we take the gold SMILES and use RDKit to compute six types of molecular attributes: molecular formula, longest carbon chain length, aromatic ring information, ring information, functional groups, and chirality. These attributes are then converted into a tagged MSR-style~\cite{jang2025structural} reasoning sequence, where each attribute is enclosed by its corresponding tag. The final supervised target is formed by concatenating this evidence sequence with the correct molecular answer in the required output format. In this way, the Generator is trained not only to output the target molecule, but also to first produce intermediate structural evidence that grounds the generation process.

\paragraph{Refiner SFT data synthesis.}
Figure~\ref{fig:data}(b) illustrates the first-stage construction of Refiner SFT data from generator-induced errors. After training the Generator, we sample 20 candidate molecules for each training description, and canonicalize the candidates and gold molecule for exact-match comparison. Non-matching candidates are treated as error cases, while their original, pre-canonicalization forms are retained as Refiner inputs. Each first-stage instance consists of the molecular description and an incorrect Generator candidate as input, with the corresponding gold molecule as the correction target. This process yields 38,891 first-stage training instances, including 23,062 from ChEBI-20 and 15,829 from PCDes.

For the second stage, we apply the Verifier to the non-matching candidates and retain those rejected by validated hard checks. Each rejected candidate is paired with its corresponding failure reason, forming an input that consists of the molecular description, rejected candidate, and verifier feedback, while the gold molecule remains the target. This process produces an additional 6,725 second-stage training instances, including 5,570 from ChEBI-20 and 1,155 from PCDes. The first stage provides broad supervision over realistic Generator errors, whereas the second stage specializes the Refiner for verification-guided correction.

\subsection{Implementation Details}

\subsubsection{Computing Infrastructure}

All experiments were conducted on a server equipped with four NVIDIA H200 GPUs, each with 140.4 GiB of memory, an Intel Xeon Platinum 8558 processor with 32 logical CPU cores, and 488 GiB of system memory. The server ran Ubuntu 22.04 LTS.

\subsubsection{LLM-Based Rule Extraction Configuration}

\begin{table}[t]
\centering
\begin{tabular}{lc}
\toprule
Parameter & Value \\
\midrule
Model & GPT-5.2 \\
Temperature & 0.5 \\
Maximum output tokens & 1,000 \\
Request timeout & 1,800 s \\
Maximum retries & 3 \\
Concurrent workers & 64 \\
\bottomrule
\end{tabular}
\caption{Configuration for LLM-based rule extraction.}
\label{tab:rule_extraction_config}
\end{table}

Table~\ref{tab:rule_extraction_config} summarizes the configuration used for LLM-based rule extraction. We use GPT-5.2 with a temperature of 0.5 and a maximum output length of 1,000 tokens. Each molecular description is processed once using the chemistry-expert system prompt and the dataset-specific rule-extraction prompt. To support large-scale extraction, requests are executed with 64 concurrent workers, a timeout of 1,800 seconds, and at most three retries. Other decoding parameters are left at API defaults.

\label{sec:app_implementation_details}
\subsubsection{Training Details}
\label{sec:app_train}

We use Intern-S1-mini~\cite{bai2025intern} as the backbone language model for molecule generation.
The generator is first fine-tuned with supervised fine-tuning (SFT) on the training split of each dataset.
After SFT, we further optimize the generator with reinforcement learning, where the policy model is initialized from the SFT checkpoint.
We implement the SFT stage with LLaMA-Factory~\cite{zheng2024llamafactory} and the RL stage with EasyR1~\cite{zheng2025easyr1}.
The final SMILES string is extracted from the content inside \texttt{\textbackslash boxed\{\}}, and the reward is computed based on the extracted answer.
The main training hyperparameters are summarized in Table~\ref{tab:training_details}. We select checkpoints based on model performance on the ChEBI-20 validation set.

\begin{table*}[t]
\centering
\small
\begin{tabular}{llll}
\toprule
Setting & Generator SFT & Generator RL & Refiner SFT (Stage 1 / Stage 2) \\
\midrule
Backbone model & Intern-S1-mini~\cite{bai2025intern} & model after SFT & Intern-S1-mini\\
Training epochs / steps & 2 epochs & 850 steps & 2.5 / 3 epochs \\
Learning rate & 1.0e-5 & 1.0e-6 & 1.0e-5 \\
Batch size & 8 & 512 & 8 \\
Optimizer & AdamW~\cite{DBLP:conf/iclr/LoshchilovH19} & AdamW & AdamW\\
Weight decay & 0.0 & 1.0e-2 & 0.0\\
Warmup ratio & 0.1 & 0.0 & 0.1 \\
Max length & 2048 & 2048+512 & 2048\\
Samples per input & -- & 8 & --\\
Precision & bf16 & AMP & bf16 \\
GPUs & 4 $\times$ H200 (140G) & 4 $\times$ H200 (140G) & 4 $\times$ H200 (140G) \\
\bottomrule
\end{tabular}
\caption{Main training hyperparameters for supervised fine-tuning and reinforcement learning.}
\label{tab:training_details}
\end{table*}

\subsubsection{Inference and Evaluation Details}
\label{sec:app_infer}

During inference, pass@1 denotes the setting in which the Generator produces one candidate for each input description. The Verifier checks the candidate and returns a failure reason when a validated constraint is violated; the rejected candidate is then refined once using the description, candidate molecule, and verifier feedback. Under pass@5, the Generator produces five candidates, which are independently verified and, when rejected, refined using their corresponding failure reasons. The refined candidates are then merged with the initially accepted candidates. From the resulting five final candidates, we select a single prediction for each input using a fixed priority order---Match, validity, Morgan FTS, RDK FTS, MACCS FTS, BLEU, and Levenshtein distance---and then compute all evaluation metrics based on these selected predictions.

In the main experimental results in Table~\ref{tab:main_results}, the results for all models are reported as pass@1. Each generated molecule is standardized and canonicalized using RDKit before evaluation. To avoid excessive RDKit parsing time caused by degenerate repetitive generations, outputs longer than 4,096 characters are not passed to RDKit-based molecular computations. They remain in the evaluation set and are counted as failures for both Validity and Match.
The main inference settings are summarized in Table~\ref{tab:inference_details}.

\begin{table}[t]
\centering
\small
\begin{tabular}{ll}
\toprule
Setting & Generator / Refiner\\
\midrule
Temperature & 0.7 \\
Top-$p$ & 0.8 \\
Top-k & 20\\
SMILES canonicalization & RDKit \\
repetition\_penalty & 1.05 \\
\bottomrule
\end{tabular}
\caption{Main inference and evaluation settings.}
\label{tab:inference_details}
\end{table}

\subsection{Evaluation Metrics}
\label{sec:metrics}
Following MSR~\cite{jang2025structural}, we evaluate generation performance using Validity, BLEU, Levenshtein distance, MACCS FTS, RDK FTS, Morgan FTS, Match, and FCD.

Let $\mathcal{D}=\{(g_i,p_i)\}_{i=1}^{N}$ denote the evaluation set, where $g_i$ and $p_i$ are the canonicalized gold and predicted SMILES for the $i$-th sample, respectively. Let $\mathrm{mol}(\cdot)$ denote RDKit parsing from SMILES to a molecular graph, and let $\mathbb{I}[\cdot]$ be the indicator function.

\begin{itemize}
    \item \textbf{Validity ($\uparrow$).}
    Validity measures the fraction of predictions that can be successfully parsed by RDKit:
    \begin{equation}
    \mathrm{Validity}
    =
    \frac{1}{N}\sum_{i=1}^{N}\mathbb{I}\!\left[\mathrm{mol}(p_i)\neq \varnothing\right].
    \end{equation}
    A higher value indicates that more predicted strings correspond to chemically valid molecules.

    \item \textbf{BLEU ($\uparrow$).}
    BLEU measures token-overlap between predictions and references. In our implementation, BLEU is computed at the \emph{character level} using corpus BLEU over the whole evaluation set:
    \begin{equation}
    \mathrm{BLEU}
    =
    \mathrm{BP}\cdot
    \exp\!\left(
    \sum_{n=1}^{4} w_n \log p_n
    \right),
    \end{equation}
    where $p_n$ is the clipped precision of character $n$-grams, $w_n=\frac{1}{4}$, and $\mathrm{BP}$ is the brevity penalty. A higher value indicates better string-level overlap with the gold SMILES.

    \item \textbf{Levenshtein ($\downarrow$).}
    Levenshtein distance measures the average string edit distance between predictions and gold answers:
    \begin{equation}
    \mathrm{Levenshtein}
    =
    \frac{1}{N}\sum_{i=1}^{N}
    \mathrm{Lev}(p_i,g_i),
    \end{equation}
    where $\mathrm{Lev}(\cdot,\cdot)$ denotes the minimum number of insertions, deletions, and substitutions needed to transform one string into the other. Lower is better.

    \item \textbf{MACCS FTS ($\uparrow$).}
    MACCS fingerprint similarity is the mean Tanimoto similarity between MACCS fingerprints of valid prediction--reference molecule pairs:
    \begin{multline}
    \mathrm{MACCS\text{-}FTS}
    =
    \frac{1}{|\mathcal{V}|}
    \sum_{i\in\mathcal{V}}
    \mathrm{Tan} \\
    \!\left(
    \phi_{\mathrm{MACCS}}(\mathrm{mol}(g_i)),
    \phi_{\mathrm{MACCS}}(\mathrm{mol}(p_i))
    \right),
    \end{multline}
    where $\mathcal{V}=\{i \mid \mathrm{mol}(g_i)\neq\varnothing,\ \mathrm{mol}(p_i)\neq\varnothing\}$, $\phi_{\mathrm{MACCS}}(\cdot)$ is the MACCS fingerprint, and $\mathrm{Tan}(\cdot,\cdot)$ is Tanimoto similarity. Higher is better.

    \item \textbf{RDK FTS ($\uparrow$).}
    RDK fingerprint similarity is defined analogously using RDKit topological fingerprints:
    \begin{multline}
    \mathrm{RDK\text{-}FTS}
    =
    \frac{1}{|\mathcal{V}|}
    \sum_{i\in\mathcal{V}}
    \mathrm{Tan} \\
    \!\left(
    \phi_{\mathrm{RDK}}(\mathrm{mol}(g_i)),
    \phi_{\mathrm{RDK}}(\mathrm{mol}(p_i))
    \right).
    \end{multline}
    Higher values indicate greater structural similarity.
    \item \textbf{Morgan FTS ($\uparrow$).}
    Morgan fingerprint similarity is computed as the mean Tanimoto similarity between Morgan fingerprints of radius $r=2$:
    \begin{multline}
    \mathrm{Morgan\text{-}FTS}
    =
    \frac{1}{|\mathcal{V}|}
    \sum_{i\in\mathcal{V}}
    \mathrm{Tan} \\
    \!\left(
    \phi_{\mathrm{Mor},r=2}(\mathrm{mol}(g_i)),
    \phi_{\mathrm{Mor},r=2}(\mathrm{mol}(p_i))
    \right).
    \end{multline}
    Higher is better.
    \item \textbf{Match ($\uparrow$).}
    Match is an exact molecular-identity metric based on InChI equality rather than raw SMILES string equality. Specifically,
    \begin{multline}
    \mathrm{Match}
    =
    \frac{1}{N}\sum_{i=1}^{N} \\
    \mathbb{I}\!\left[
    \mathrm{InChI}(\mathrm{mol}(p_i))
    =
    \mathrm{InChI}(\mathrm{mol}(g_i))
    \right].
    \end{multline}
    Invalid predictions are counted as mismatches. A higher value indicates that more predictions correspond to exactly the same molecule as the gold target.

    \item \textbf{FCD ($\downarrow$).}
    Fr\'echet ChemNet Distance (FCD) is a dataset-level distributional metric computed on canonicalized SMILES. Let $\mu_g,\Sigma_g$ and $\mu_p,\Sigma_p$ be the empirical mean and covariance of ChemNet features extracted from the gold and predicted molecule sets, respectively. Then
    \begin{multline}
    \mathrm{FCD}
    =
    \|\mu_g-\mu_p\|_2^2
    + \\
    \mathrm{Tr}\!\left(
    \Sigma_g+\Sigma_p
    -
    2(\Sigma_g\Sigma_p)^{1/2}
    \right).
    \end{multline}
    Lower values indicate that the predicted molecule distribution is closer to the gold distribution.
\end{itemize}

\paragraph{Multi-sample evaluation.}
For multi-sample settings, we first obtain multiple candidates for each input and then select a single final candidate per sample for table-ready evaluation. In the provided implementation, this selection follows a fixed priority:
\begin{multline}
\texttt{Match}
\;>\;
\texttt{Validity}
\;>\;
\texttt{Morgan}
\;>\;
\texttt{RDK} \\
\;>\;
\texttt{MACCS}
\;>\;
\texttt{BLEU}
\;>\;
-\texttt{Levenshtein}.
\end{multline}
That is, among all candidates for a sample, we keep the one with the highest tuple under this lexicographic ordering, and then compute the dataset-level metrics on the resulting set of selected predictions.

\begin{figure*}[!t]
\centering
\includegraphics[width=\textwidth]{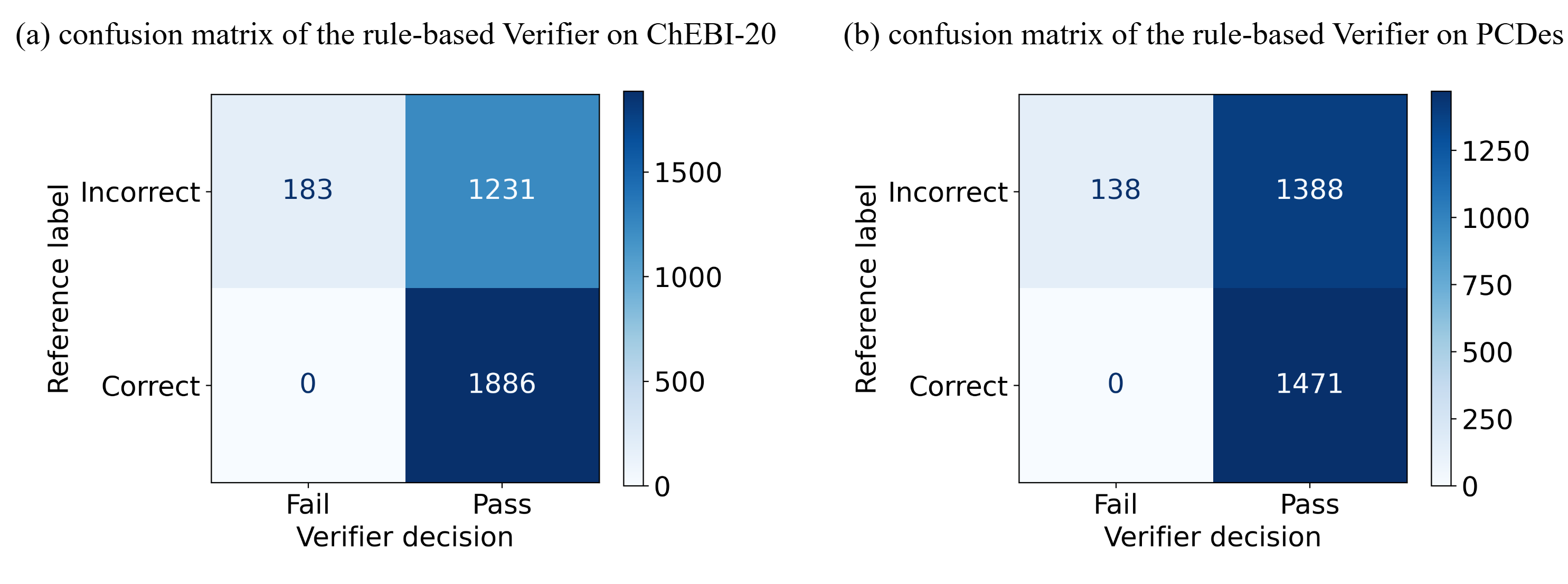}

\caption{
Confusion matrices of the rule-based MolGVR Verifier on ChEBI-20 and PCDes.
Under the expert-validated hard-check rules, the Verifier produces no false rejections on either dataset while identifying 183 and 138 rule-violating candidates on ChEBI-20 and PCDes, respectively, demonstrating conservative rejection behavior.
}
\label{fig:verifier_confusion}
\end{figure*}

\section{Evaluation of the Rule-based Verifier}
\label{sec:app_eval_verifier}

Figure~\ref{fig:verifier_confusion} shows the decision behavior of the rule-based MolGVR Verifier under the single-generation setting, where the Generator produces one candidate for each input and the Verifier then makes a pass/fail decision.
The Verifier is conservative: it produces no false rejections on either dataset, indicating that correct molecules are not mistakenly filtered out by the current rules.
At the same time, it rejects 183/1414 incorrect candidates on ChEBI-20 and 138/1526 on PCDes, indicating that the current set of checkable constraints covers only a subset of molecule-level errors.
This behavior reflects a precision-oriented trade-off: since description-derived constraints can be difficult to verify reliably with current rule-based RDKit checks, we only use expert-validated hard-check rules for rejection.
This conservative strategy reduces the risk of sending correct molecules to the Refiner while providing reliable failure feedback for targeted correction.
\section{Inference time of MolGVR}
\label{sec:app_inference_time}

To further evaluate the computational overhead introduced by the proposed
generate--verify--refine pipeline, we report the wall-clock inference time on
the ChEBI-20 test set. Both MolGVR and ChemDFM-v1.5-8B are deployed with vLLM
on a single NVIDIA H200 GPU with 140GB memory. For both models, we use the same
main decoding configuration: batch size 64, tensor parallel size 1, temperature
0.7, top-$p$ 0.8, top-$k$ 20, and repetition penalty 1.05. For the @1 setting,
we sample one candidate per input, while for the @5 setting, we sample five
candidates per input.
The rule extraction results are precomputed and cached, and we report the online inference cost after rule extraction.
As shown in Table~\ref{tab:inference_time}, MolGVR introduces moderate
additional inference cost compared with ChemDFM-v1.5-8B. Specifically, MolGVR@1
takes 39.22 seconds longer than
ChemDFM@1. Under the @5 setting, MolGVR takes 79.33 seconds longer than ChemDFM@5. This overhead mainly
comes from the additional verifier and refiner stages. Nevertheless, the runtime
increase remains relatively small, while MolGVR achieves a stronger exact-match
performance and better description-level molecular correctness. These results
suggest that the verification-and-refinement mechanism improves generation
quality with a modest and practical inference-time overhead.

\begin{table}[t]
\centering
\small
\begin{tabular}{lcc}
\toprule
Setting & ChemDFM-v1.5-8B & MolGVR \\
\midrule
@1
& 248.51
& 287.73 (221.60 + 1.86 + 64.27) \\
@5
& 473.35
& 552.68 (350.17 + 7.93 + 194.58) \\
\bottomrule
\end{tabular}
\caption{
Inference-time comparison on the ChEBI-20 test set. All times are reported in
seconds. For MolGVR, the three addends in parentheses
denote the time of generator inference, verifier execution, and refiner
inference, respectively.
}
\label{tab:inference_time}
\end{table}

\section{More Examples}
\subsection{ChEBI-20}
\label{sec:more_exp_chebi20}
System prompt for generator: You are an expert in generating molecules from text descriptions. You will receive a text description as input, and you should first reason about six molecular properties within <think> tags---namely the molecular formula, the longest carbon chain length, the number of aromatic rings, ring compounds, functional groups, and chirality---and then predict the molecule's SMILES in \texttt{\textbackslash \textbackslash}boxed\{\} format. The output must follow this format: "<think> <mol\_formula> ... </mol\_formula> <longest\_carbon\_chain> ... </longest\_carbon\_chain> <aromatic\_ring> ... </aromatic\_ring> <ring> ... rings of size ... </ring> <fun\_group> ... </fun\_group> <chiral> ... with S; ... with R </chiral> </think> \texttt{\textbackslash \textbackslash}boxed\{final smiles\}" format.

\noindent System prompt for Refiner: You are a molecular SMILES refinement model. Your task is to correct an incorrect candidate SMILES using the molecule description and verifier feedback. The molecule description is the primary source of truth. The verifier feedback identifies the types of structural constraints violated by the candidate and the exact text requirements from the description that are not satisfied. Correct all reported constraint violations. The incorrect candidate SMILES is only a noisy hint and may be partially wrong or entirely wrong. Preserve candidate substructures only when they are consistent with both the description and the verifier feedback. Return only the final answer in the following format: The refined SMILES is \texttt{\textbackslash \textbackslash}boxed{ correct smiles }.

\subsubsection{Case 1}

\textbf{Generator input:}

\noindent<tbdd>\texttt{\textbackslash n}The molecule is a polyazaalkane consisting of a 13-membered straight chain alkane with aza-groups at the 1-, 5-, 9- and 13-positions. It is a polyazaalkane and a tetramine. It is a conjugate base of a 3,3,3-tetraminium(4+).\\
\textbf{Generator output:}

\begin{myverb}
<mol_formula> C9H24N4 </mol_formula>\t<longest_carbon_chain> 3 </longest_carbon_chain>\t<aromatic_ring> 0 </aromatic_ring>\t<ring> 0 </ring>\t<fun\_group> amine, organic, primary, secondary </fun_group>\t<chiral> 0: 0 with S; 0 with R </chiral>  </think> \nBased on the reasoning above, I suppose the final SMILES is \\boxed{C(CN)CNCCCN}
\end{myverb}
\textbf{Verify rule:}
\begin{myverb}
[
  {"category": "functional_groups", "text_span": "tetramine", "normalized": "amine"},
  {"category": "elements_or_atom_types", "text_span": "aza-groups", "normalized": "nitrogen"},
  {"category": "element_count_requirements", "text_span": "aza-groups at the 1-, 5-, 9- and 13-positions", "normalized": "nitrogen", "count": 4},
  {"category": "group_count_requirements", "text_span": "tetramine", "normalized": "amino_group", "count": 4}
]
\end{myverb}
\textbf{Refiner input:}
\begin{myverb}
Task: Refine the incorrect candidate SMILES into the correct target molecule.

Molecule description:
The molecule is a polyazaalkane consisting of a 13-membered straight chain alkane with aza-groups at the 1-, 5-, 9- and 13-positions. It is a polyazaalkane and a tetramine. It is a conjugate base of a 3,3,3-tetraminium(4+).

Incorrect candidate SMILES:
C(CN)CNCCCN

Verifier feedback:
1. Error category: element_count_requirements
   Unmet text requirement: "aza-groups at the 1-, 5-, 9- and 13-positions"

2. Error category: group_count_requirements
   Unmet text requirement: "tetramine"

Correct all listed violations and return the target molecule.

Output format:
The refined SMILES is \\boxed{ correct smiles }
\end{myverb}
\textbf{Refiner output:}
\begin{myverb}
</think> \n\nThe refined SMILES is \\boxed{C(CN)CNCCCNCCCN}
\end{myverb}

\subsubsection{Case 2}

\textbf{Generator input:}

\noindent<tbdd>\texttt{\textbackslash n}The molecule is a hydroxy fatty acid anion that is the conjugate base of omega-hydroxy-15-methyl-palmitic acid, obtained by deprotonation of the carboxy group; major species at pH 7.3. It is a branched-chain saturated fatty acid anion, a long-chain fatty acid anion, a methyl-branched fatty acid anion and an omega-hydroxy fatty acid anion. It derives from an isoheptadecanoate. It is a conjugate base of an omega-hydroxy-15-methylpalmitic acid.\\
\textbf{Generator output:}

\begin{myverb}
<mol_formula> C17H33O3- </mol_formula>\t<longest_carbon_chain> 16 </longest_carbon_chain>\t<aromatic_ring> 0 </aromatic_ring>\t<ring> 0 </ring>\t<fun_group> alcohol, carboxylate, organic </fun_group>\t<chiral> 0: 0 with S; 0 with R </chiral>  </think> \nBased on the reasoning above, I suppose the final SMILES is \\boxed{CC(C)CCCCCCCCCCCCCC(=O)[O-]}
\end{myverb}
\textbf{Verify rule:}
\begin{myverb}
[
  {"category": "functional_groups", "text_span": "hydroxy fatty acid anion", "normalized": "carboxylate"},
  {"category": "functional_groups", "text_span": "omega-hydroxy fatty acid anion", "normalized": "alcohol"},
  {"category": "functional_groups", "text_span": "deprotonation of the carboxy group", "normalized": "carboxylate"}
]
\end{myverb}
\textbf{Refiner input:}
\begin{myverb}
Task: Refine the incorrect candidate SMILES into the correct target molecule.

Molecule description:
The molecule is a hydroxy fatty acid anion that is the conjugate base of omega-hydroxy-15-methyl-palmitic acid, obtained by deprotonation of the carboxy group; major species at pH 7.3. It is a branched-chain saturated fatty acid anion, a long-chain fatty acid anion, a methyl-branched fatty acid anion and an omega-hydroxy fatty acid anion. It derives from an isoheptadecanoate. It is a conjugate base of an omega-hydroxy-15-methylpalmitic acid.

Incorrect candidate SMILES:
CC(C)CCCCCCCCCCCCCC(=O)[O-]

Verifier feedback:
1. Error category: functional_groups
   Unmet text requirement: "omega-hydroxy fatty acid anion"

Correct all listed violations and return the target molecule.

Output format:
The refined SMILES is \\boxed{ correct smiles }
\end{myverb}
\textbf{Refiner output:}
\begin{myverb}
</think> \n\nThe refined SMILES is \\boxed{CC(CCCCCCCCCCCCCC(=O)[O-])CO}
\end{myverb}

\subsubsection{Case 3}

\textbf{Generator input:}

\noindent<tbdd>\texttt{\textbackslash n}The molecule is a 2,3-dihydro-3-hydroxyanthranilic acid in which both stereocentres have S-configuration. It is an enantiomer of a (2R,3R)-2,3-dihydro-3-hydroxyanthranilic acid. It is a tautomer of a (2S,3S)-2,3-dihydro-3-hydroxyanthranilic acid zwitterion.\\
\textbf{Generator output:}

\begin{myverb}
\\boxed{C1=C[C@@H]([C@@H](C(=C1)C(=O)O)N)O}
\end{myverb}
\textbf{Verify rule:}
\begin{myverb}
[
  {"category": "functional_groups", "text_span": "hydroxyanthranilic acid", "normalized": "carboxylic_acid"},
  {"category": "functional_groups", "text_span": "hydroxyanthranilic acid", "normalized": "alcohol"},
  {"category": "stereochemistry_requirements", "text_span": "both stereocentres have S-configuration", "normalized": "s_center", "count": 2}
]
\end{myverb}
\textbf{Refiner input:}
\begin{myverb}
Task: Refine the incorrect candidate SMILES into the correct target molecule.

Molecule description:
The molecule is a 2,3-dihydro-3-hydroxyanthranilic acid in which both stereocentres have S-configuration. It is an enantiomer of a (2R,3R)-2,3-dihydro-3-hydroxyanthranilic acid. It is a tautomer of a (2S,3S)-2,3-dihydro-3-hydroxyanthranilic acid zwitterion.

Incorrect candidate SMILES:
C1=C[C@@H]([C@@H](C(=C1)C(=O)O)N)O

Verifier feedback:
1. Error category: stereochemistry_requirements
   Unmet text requirement: "both stereocentres have S-configuration"

Correct all listed violations and return the target molecule.

Output format:
The refined SMILES is \\boxed{ correct smiles }
\end{myverb}
\textbf{Refiner output:}
\begin{myverb}
</think> \n\nThe refined SMILES is \\boxed{C1=C[C@@H]([C@H](C(=C1)C(=O)O)N)O}
\end{myverb}

\subsection{PCDes}
\label{sec:more_exp_pcdes}

\subsubsection{Case 1}

\textbf{Generator input:}

\noindent<tbdd>\texttt{\textbackslash n}it is an aminoalkylpyridine that is pyridine substituted by a ethanamino group at position 2. It has a role as a metabolite and a histamine agonist. It is a primary amine and an aminoalkylpyridine. It derives from a pyridine.\\
\textbf{Generator output:}

\begin{myverb}
\\boxed{CCNC1=CC=CC=N1}
\end{myverb}
\textbf{Verify rule:}
\begin{myverb}
[
  {"category": "functional_groups", "text_span": "primary amine", "normalized": "amine"},
  {"category": "core_scaffolds_or_ring_systems", "text_span": "aminoalkylpyridine", "normalized": "pyridine"},
  {"category": "elements_or_atom_types", "text_span": "primary amine", "normalized": "nitrogen"}
]
\end{myverb}
\textbf{Refiner input:}
\begin{myverb}
Task: Refine the incorrect candidate SMILES into the correct target molecule.

Molecule description:
it is an aminoalkylpyridine that is pyridine substituted by a ethanamino group at position 2. It has a role as a metabolite and a histamine agonist. It is a primary amine and an aminoalkylpyridine. It derives from a pyridine.

Incorrect candidate SMILES:
CCNC1=CC=CC=N1

Verifier feedback:
1. Error category: functional_groups
   Unmet text requirement: "primary amine"

Correct all listed violations and return the target molecule.

Output format:
The refined SMILES is \\boxed{ correct smiles }
\end{myverb}
\textbf{Refiner output:}
\begin{myverb}
</think> \n\nThe refined SMILES is \\boxed{C1=CC=NC(=C1)CCN}
\end{myverb}

\subsubsection{Case 2}

\textbf{Generator input:}

\noindent<tbdd>\texttt{\textbackslash n}it is the prodrug of fluphenazine, an antipsychotic drug used for the symptomatic management of psychosis in patients with schizophrenia. It has a role as a phenothiazine antipsychotic drug, a prodrug and a dopaminergic antagonist. It is a member of phenothiazines, a decanoate ester, an organofluorine compound and a N-alkylpiperazine. It derives from a fluphenazine. it is the decanoate salt form of fluphenazine, a phenothiazine with antipsychotic activity. it exerts its actions by blocking postsynaptic dopamine D2 receptors in the limbic, cortical system and basal ganglia. This prevents the actions of dopamine, thereby reducing the hallucinations and delusions that are associated with schizophrenia.\\
\textbf{Generator output:}

\begin{myverb}
<mol_formula> C30H41F3N2O2S </mol_formula>\t<longest_carbon_chain> 10 </longest_carbon_chain>\t<aromatic_ring> 2 </aromatic_ring>\t<ring> 3 rings of size 6 </ring>\t<fun_group> amine, aromatic, ester, ether, fluoroalkane, haloalkane, organic, sulfide, tertiary </fun_group>\t<chiral> 0: 0 with S; 0 with R </chiral>  </think> \n Based on the reasoning above, I suppose the final SMILES is \\boxed{CCCCCCCCCC(=O)N1CCN(CC1)CCCN2C3=CC=CC=C3SC4=C2C=C(C=C4)C(F)(F)F}
\end{myverb}
\textbf{Verify rule:}
\begin{myverb}
[
  {"category": "functional_groups", "text_span": "a decanoate ester", "normalized": "ester"},
  {"category": "elements_or_atom_types", "text_span": "an organofluorine compound", "normalized": "fluorine"}
]
\end{myverb}
\textbf{Refiner input:}
\begin{myverb}
Task: Refine the incorrect candidate SMILES into the correct target molecule.

Molecule description:
it is the prodrug of fluphenazine, an antipsychotic drug used for the symptomatic management of psychosis in patients with schizophrenia. It has a role as a phenothiazine antipsychotic drug, a prodrug and a dopaminergic antagonist. It is a member of phenothiazines, a decanoate ester, an organofluorine compound and a N-alkylpiperazine. It derives from a fluphenazine. it is the decanoate salt form of fluphenazine, a phenothiazine with antipsychotic activity. it exerts its actions by blocking postsynaptic dopamine D2 receptors in the limbic, cortical system and basal ganglia. This prevents the actions of dopamine, thereby reducing the hallucinations and delusions that are associated with schizophrenia.

Incorrect candidate SMILES:
CCCCCCCCCC(=O)N1CCN(CC1)CCCN2C3=CC=CC=C3SC4=C2C=C(C=C4)C(F)(F)F

Verifier feedback:
1. Error category: functional_groups
   Unmet text requirement: "a decanoate ester"

Correct all listed violations and return the target molecule.

Output format:
The refined SMILES is \\boxed{ correct smiles }
\end{myverb}
\textbf{Refiner output:}
\begin{myverb}
</think> \n\nThe refined SMILES is \\boxed{CCCCCCCCCC(=O)OCCN1CCN(CC1)CCCN2C3=CC=CC=C3SC4=C2C=C(C=C4)C(F)(F)F}
\end{myverb}

\subsubsection{Case 3}

\textbf{Generator input:}

\noindent<tbdd>\texttt{\textbackslash n}it appears as white to off-white crystals or powder with a fishy odor. Initial pH (2\% aqueous solution) 3.0-4.0. (NTP, 1992) it is the hydrochloride salt of mechlorethamine. It has a role as an antineoplastic agent. It contains a mechlorethamine. Mechlorethamine Hydrochloride is the hydrochloride salt of mechlorethamine, a nitrogen mustard and an analogue of sulfur mustard, with antineoplastic and immunosuppressive activities. Mechlorethamine is metabolized to an unstable, highly reactive ethyleniminium intermediate that alkylates DNA, particularly the 7 nitrogen of guanine residues, resulting in DNA base pair mismatching, DNA interstrand crosslinking, the inhibition of DNA repair and synthesis, cell-cycle arrest, and apoptosis. This agent also exhibits lympholytic properties.\\
\textbf{Generator output:}

\begin{myverb}
\\boxed{CN(CCCl)CCCl}
\end{myverb}
\textbf{Verify rule:}
\begin{myverb}
[
  {"category": "elements_or_atom_types", "text_span": "a nitrogen mustard", "normalized": "nitrogen"},
  {"category": "salt_or_charge_state", "text_span": "the hydrochloride salt of mechlorethamine", "normalized": "hydrochloride_salt"}
]
\end{myverb}
\textbf{Refiner input:}
\begin{myverb}
Task: Refine the incorrect candidate SMILES into the correct target molecule.

Molecule description:
it appears as white to off-white crystals or powder with a fishy odor. Initial pH (2\% aqueous solution) 3.0-4.0. (NTP, 1992) it is the hydrochloride salt of mechlorethamine. It has a role as an antineoplastic agent. It contains a mechlorethamine. Mechlorethamine Hydrochloride is the hydrochloride salt of mechlorethamine, a nitrogen mustard and an analogue of sulfur mustard, with antineoplastic and immunosuppressive activities. Mechlorethamine is metabolized to an unstable, highly reactive ethyleniminium intermediate that alkylates DNA, particularly the 7 nitrogen of guanine residues, resulting in DNA base pair mismatching, DNA interstrand crosslinking, the inhibition of DNA repair and synthesis, cell-cycle arrest, and apoptosis. This agent also exhibits lympholytic properties.

Incorrect candidate SMILES:
CN(CCCl)CCCl

Verifier feedback:
1. Error category: salt_or_charge_state
   Unmet text requirement: "the hydrochloride salt of mechlorethamine"

Correct all listed violations and return the target molecule.

Output format:
The refined SMILES is \\boxed{ correct smiles }
\end{myverb}
\textbf{Refiner output:}
\begin{myverb}
</think> \n\nThe refined SMILES is \\boxed{CN(CCCl)CCCl.Cl}
\end{myverb}

\end{document}